\documentclass[twocolumn, 10pt]{article}

\usepackage{cite}
\usepackage{amsmath,amssymb,amsfonts}
\usepackage{textcomp}
\usepackage{xcolor}
\usepackage{microtype}
\usepackage{graphicx}
\usepackage{caption}
\usepackage{subcaption}
\usepackage{booktabs}
\usepackage[margin={2cm,2cm}]{geometry}
\usepackage{hyperref}

\usepackage{xargs}
\usepackage{xspace}
\usepackage{outlines}
\usepackage{diagbox}
\usepackage{pgfgantt}
\usepackage{cleveref}

\begin{document}

\title{MG-GCN: Scalable Multi-GPU GCN Training Framework}

\author{Muhammed Fatih Bal{\i}n\thanks{Equal contribution}\textsuperscript{ ,}\thanks{School of Computational Science and Engineering, Georgia Institute of Technology, Atlanta, Georgia, USA}
\and
Kaan Sancak\footnotemark[1]\textsuperscript{ ,}\footnotemark[2]
\and
\"{U}mit V. \c{C}ataly\"{u}rek \footnotemark[2]\textsuperscript{ ,}\thanks{Amazon Web Services. This publication describes work performed at the Georgia Institute of Technology and is not associated with Amazon.}}

\date{\{balin, kaan, umit\}@gatech.edu}

\maketitle
\begin{abstract}
    Full batch training of Graph Convolutional Network (GCN) models
    is not feasible on a single GPU for large graphs containing tens of millions
    of vertices or more. Recent work has shown that, for the graphs used in the
    machine learning community, communication becomes a bottleneck and scaling
    is blocked outside of the single machine regime. Thus, we propose MG-GCN,
    a multi-GPU GCN training framework taking advantage of the high-speed
    communication links between the GPUs present in multi-GPU systems.
    MG-GCN employs multiple High-Performance Computing optimizations,
    including efficient re-use of memory buffers to reduce
    the memory footprint of training GNN models, as well as communication and
    computation overlap. These optimizations enable execution on larger datasets,
    that generally do not fit into memory of a single GPU in state-of-the-art implementations.
    Furthermore, they contribute to achieve superior speedup compared to the
    state-of-the-art. For example, MG-GCN achieves super-linear speedup with respect
    to DGL, on the Reddit graph on both DGX-1 (V100) and DGX-A100.
\end{abstract}

\section{Introduction}

Graphs are essential data-structures that can represent a variety of information, therefore they
surface in many different contexts and disciplines. The Graph Convolutional Network (GCN) model is a
type of Graph Neural Network (GNN) which is a
very powerful graph embedding method for semi-supervised learning to solve graph representation
learning problems~\cite{GNN, gcn}. GNNs take advantage of the connectivity information presented in
the graph, thus they provide flexibility and greater applicability compared to CNN models where the neighborhood structure of nodes is fixed, hence the model is
more restricted. The common use cases of GNN models include \emph{node prediction}~\cite{gcn} which predicts
the properties of certain vertices, \emph{graph prediction}~\cite{zhang2018end} which predicts the properties of the
whole graph, and \emph{link prediction}~\cite{zhang2018link} which predicts whether there is an edge exists between
two nodes. In this work, we will focus on node prediction, but our methods are extendable
to graph and link prediction as well.

While training GNNs, the memory requirement for large graphs can exceed the memory capacity of a
single accelerator. \emph{Mini-batch} training is a common technique to overcome this problem 
to reduce the working set by creating a mini-batch of vertex samples to
train the model. Consequently, it reduces the memory requirement during training.
However, mini-batch training might lead to important problems. First, starting from the
mini-batch nodes, it is possible to reach almost every single node in the graph in just a few hops,
also known as neighborhood explosion phenomenon, which increases the work performed during a single
epoch exponentially. Second, it has been shown that mini-batch training can lead to lower accuracy
compared to full-batch training~\cite{ROC}. In this work, we focus on full-batch training on
multi-GPU systems.

A major challenge to full-batch GCN training is their parallelization and scalability.
The challenge stems mainly from the irregular structure of the graph which leads to load imbalance
and communication cost when training on multiple GPUs. GCN has many underlying
kernels, however, one of the most time consuming part is the Sparse Matrix-Dense
Matrix Multiplications (SpMM). Alternative solutions are proposed to
improve the performance of SpMM, such as reordering and better suited graph storage schemes and
computation kernels~\cite{spmm}.

\begin{figure*}[ht!]
    \centering
    \includegraphics[width=1.8\columnwidth]{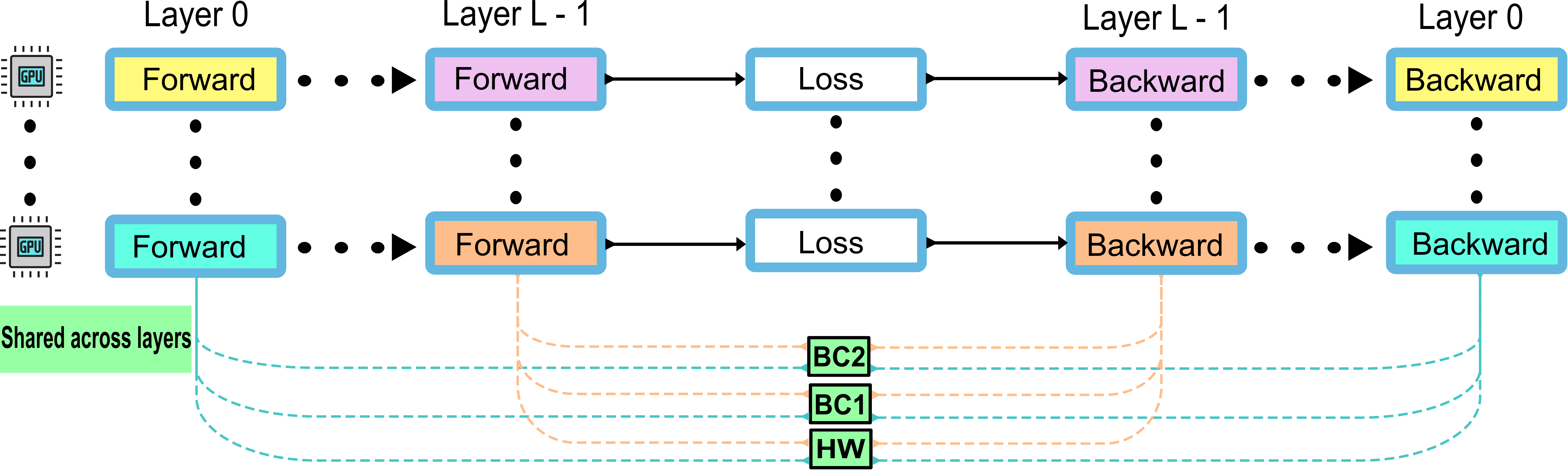}
    \caption{\textbf{Computation diagram} of an $L$-layer GCN model with shared buffers across layers. $BC1$: broadcast buffer, $BC2$: broadcast buffer for overlapping, $HW$: temporary result buffer between SpMM and GeMM.}
    \label{fig:computation_diagram}
\end{figure*}

Most of the existing systems, such as Deep Graph Learning Library (DGL), lack the support for
multi-GPU training~\cite{Wang2019DeepGL}. One needs to
implement the parallelism manually while using DGL.
DistDGL is an extension
of DGL that enables multi-GPU training, however, it does not provide full-batch training,
rather it uses mini-batch training~\cite{DistDGL}.
Recently, ROC~\cite{ROC} has been proposed and it supports automatic multi-GPU
GCN full-batch training on a single machine, and scales up to multiple machines.
CAGNET~\cite{CAGNET} builds on top of ROC by providing distributed
algorithms with different communication patterns. In their work, authors
investigate different partitioning strategies to reduce the communication cost
and scale up-to hundreds of GPUs.
However, their results show that none of the proposed algorithms
is able to scale beyond a single node (4 GPUs), primarily due to restricted bandwidth of the
available interconnect between nodes in the cluster.

In this work, we provide a framework for training GCNs on multiple GPUs that takes advantage
of the high-speed communication links present in today's multi-GPU systems~\cite{dgx}.
We address the load imbalance problem by using simple random permutation strategy,
and hide the communication by overlapping it with computation. Moreover, we carefully examine the dependency scheme of
the buffers used during training and investigate ways to reduce
memory requirement for GCN models to fit larger datasets into our target
machines. Our optimization techniques are generalizable and can be applied to
other frameworks but for reproducibility, we also release our customizable
implementation of MG-GCN, as an open-source library.

\section{Background}
\label{sec:background}

The inputs for a GCN $f_A(X)$ are the feature matrix $X
\in \mathbb{R}^{n \times d}$ and the adjacency matrix $A \in \mathbb{R}^{n
\times n}$, when there are $n$ input instances and each input instance has $d$
dimensional feature vectors. GCNs are useful when the
input instances come equipped with a relation, which is represented in matrix
format as $A$. Typically, input instances have features along them which make up
$X$. To do learning on such a dataset, one can ignore $A$ and fit a model
that treats each input instance independently using a multi-layer perceptron. In
contrast, GCNs utilize $A$, and instead of processing
each input instance separately, it processes an input instance together with its
$k$-hop neighborhood. Having access to an instance's neighborhood increases the
expressiveness of the model, hence aids performance immensely.
As an example, consider guessing which movies an individual
would like to watch.
It might prove to be a hard task if we have access to a single individual.
However, if we consider a group of individuals that are related to the person of
interest, then the prediction task becomes much simpler as individual variance
vanishes whereas group difference becomes more visible when one looks at whole
groups at once. This is why GCN often perform much
better compared to simple multi-layer perceptron models that do not take into
account the relations of instances~\cite{gcn}.

The simplest variant of a GCN $f_A(H)$ with a
single layer can be represented as
\begin{align}
    \label{eq:gcn_layer}
    f_A(H) &= \sigma(\hat{A}^THW) \\
    \hat{A}_{uv} &= \frac{A_{uv}}{\sum_{w \in \mathcal{N}_i(v)} A_{wv}}
\end{align}
where $\mathcal{N}_i(v)$ is the set of in-edges for vertex $v$ and $\sigma$ is
an element-wise nonlinear activation function, ReLU~\cite{relu} in our case. Using $f_A$, we can construct
deeper GCNs as follows for any number of layers $L$:
\begin{align}
    H^{(0)} &= X \\
    & \vdots \nonumber \\
    H^{(L)} &= f_A(H^{(L-1)})
\end{align}
As depicted in Figure~\ref{fig:computation_diagram}, $L$-layer GCN model
training is composed to $L$ forwarded passes followed by $L$ backward passes.
More specifically, given input matrix $H$, the operations in the
forward pass of a single GCN layer can be broken down as follows:
\begin{align}
    \label{eq:forward_pass_gemm}
    HW &= H * W \\
    \label{eq:forward_pass_spmm}
    AHW &= \hat{A}^T * HW \\
    \label{eq:forward_pass_activation}
    H' &= \sigma(AHW)
\end{align}
where  $*$ denotes the matrix multiplication operation.
Similarly given the gradient from the next layer $H_G'$, the backward layer can be broken down as follows:
\begin{align}
    \label{eq:backward_pass_activation}
    AHW_G &= \sigma'(H_G', AHW) \\
    \label{eq:backward_pass_spmm}
    HW_G &= \hat{A} * AHW_G \\
    \label{eq:backward_pass_gemm}
    W_G &= HW_G^T * H \\
    \label{eq:backward_pass_gemm1}
    H_G &= HW_G * W^T
\end{align}
where we use subscript $G$ in $U_G$, to denote the derivative of $U$ with respect to
the loss function.

As we will experimentally verify in Section~\ref{sec:runtime_decomposition},
at the core of these GCN computations,
there are two operations which are computationally the most expensive:
1)~Sparse Matrix-dense
Matrix multiplications (SpMM) in $\hat{A}^T*HW$ and $\hat{A} * AHW_G$, and
2)~(General) dense Matrix-Matrix multiplication (GeMM) operations, in $HW$,
$HW_G * W^T$ and $HW_G^T * H$.
For efficient parallel and distributed
execution, one needs to pay attention to these two operations.

\section{Related Work}

The growing size and scale of data encouraged many researchers to develop
parallel/distributed algorithms and systems for Deep Neural
Networks~\cite{lenet,ben2019demystifying}. Broady, DNN parallelism
can be generalized under 3 categories: data parallelism, model parallelism,
and pipelining. Data parallelism can be further divided into 2 categories.
Mini-batch parallelism creates batches from the dataset by using sampling methods,
and then partitions the batches among compute
resources~\cite{goyal2017accurate, ginsburg2018large},
while Coarse- and Fine-Grained or full-batch
parallelism divides the dataset among the compute
resources~\cite{zinkevich2010parallelized, Zhang2014}. On the other hand,
model parallelism divides the model itself, and partitions the work depending
on the neurons in each layer~\cite{coates2013deep, ericson2017performance}.
Alternatively, pipelining can be achieved in
two ways. Either overlapping the computations between consecutive layers,
or partitioning the model according to its depth, and dividing layers among
processors~\cite{scalablestacking, tensorflow, chen2015mxnet, pytorch}.
Also, there has been hybrid approaches that combine
multiple parallelism schemes~\cite{krizhevsky2014one}.

While alternative methods exists, most of the research on GNN parallelism is focused on
data parallelism, since the models are relatively simple compared to the traditional DNN models.
Similar to DNNs, data parallelism can be achieved in two ways. Mini-batch, or sampling, based
approaches create batches via neighborhood sampling~\cite{chen2018fastgcn, chiang2019cluster}.
After batches are created, they are assigned to CPUs or GPUs. However, in the case of graphs,
mini-batching might result in neighborhood explosion in just few hops, increasing the work
performed in an epoch exponentially. Alternatively, to avoid
the computation waste, one can apply full-batch parallelism where the parallelism
achieved by distributing the workload among CPUs/GPUs while
keeping execution order of the layers identical to the sequential method~\cite{CAGNET, md2021distgnn}.
In full-batch training, the model takes the whole graph and the corresponding
features as input, and to achieve any parallelism one has to apply ideas similar
to the model parallelism in general DNNs since the work of individual layers
has to be partitioned. In this work, we focus on this aspect of GNN model training.

Most of the CPU-based systems are focused on mini-batching based methods. 
AliGraph is a comprehensive distributed GNN training framework that
provides aggregators and operators for various GNN models~\cite{Aligraph}.
\mbox{AliGraph} enables 4 different partitioning algorithms: METIS, Vertex cut
\& Edge Cut, 2D partitioning, and Streaming-style partitioning. However, it
neither provides much details on the subject nor includes any scaling
experiments. DistDGL~\cite{DistDGL} is a framework based on DGL that uses METIS
partitioning~\cite{metis}. It keeps vertex and edge features in a
distributed key-value store, which can be queried during the training. DistDGL
shows scaling results on the largest available benchmark datasets. However, none
of these frameworks provides support for training on the full-graph.
DistGNN~\cite{md2021distgnn} is a scalable distributed training framework for
large-scale GNNs that is an extension of DGL. Unlike other frameworks, DistGNN
trains the models on the full graph. It uses a vertex-cut partitioning
called Libra~\cite{Libra}, and shows substantial scaling on largest available benchmark
datasets. However, as we will show in the later sections, by applying extensive memory
optimizations, we are able to fit some of the largest datasets using only 1 to 8
GPUs, while achieving 12.5x faster runtimes than DistGNN's best performance
which is achieved with up to 128 sockets.

There has been various frameworks and algorithms proposed for training GNNs on
GPUs. Deep Graph Library (DGL)~\cite{Wang2019DeepGL} is a well-known library for
implementing general GNN models. DGL provides the API for sparse matrix operations and
sampling functions to implement various GNN models efficiently. Moreover, it can use
Tensorflow~\cite{tensorflow}, PyTorch~\cite{pytorch} or
MXNet~\cite{chen2015mxnet} as backends for wide adoption.
NeuGraph~\cite{neugraph} is a single node multi-GPU mini-batch GNN training
framework. NeuGraph introduces a programming model for GNN computations that is
similar to vertex-centric programming model~\cite{powergraph}.
ROC~\cite{ROC} is a distributed multi-GPU GNN training framework utilizing
graph partitioning via an online regression model and it proposes memory management
optimizations for transfers of data between the CPU and GPU.
ROC shows scalability on some of the available benchmark datasets such as Reddit
and Amazon, and also it is able to do full-batch training of more complex models
and achieve higher accuracy compared to sampling approaches. However,
we are not able to compare with ROC, since they do not provide multi-gpu
results in their work and the available code does not work as expected.
CAGNET~\cite{CAGNET}, inspired by the SUMMA algorithm~\cite{Geijn1997SUMMASU},
implements 1D, 1.5D, 2D and 3D partitioning strategies for full-batch training
in order to reduce the communication cost.
Additionally, authors provide a complexity analysis for each strategy.
However, CAGNET fails to scale beyond a single node (4 GPUs) in terms of runtime
performance due to the available bandwidth and the intra and inter communication
topologies. Moreover, CAGNET does not have an effort to reuse memory buffers,
and it relies on PyTorch and PyTorch Geometric libraries~\cite{pyg}.
As we will show in the later sections, by adapting extensive memory optimizations, we
are able to fit much larger graphs into our target machines.

\section{MG-GCN}

Looking at a single layer of a GCN model particularly, we can express it via the following:
\begin{equation}
    H^{(l + 1)} = f_A(H^{(l)}) = \sigma (\hat{A}^TH^{(l)} W^{(l)})
\end{equation}
where matrices $A$, $H$, and $W$ are defined in Section~\ref{sec:background}. That is, one layer of
GCN consists of two main operations. For the forward propagation, first, we need
to perform a Generalized Matrix Matrix Multiplication (GeMM) between the dense
matrices H and W, then we need to perform a Sparse Matrix-Dense Matrix
Multiplication (SpMM) between the transpose sparse matrix A, and the resulting
matrix of GeMM. Later the result of SpMM is fed into a nonlinear activation
function. In the backward pass, same operations are performed with the
non-transposed normalized adjacency matrix $\hat{A}$. In the rest of this section,
we will focus on the forward pass and we refer to $\hat{A}^T$ simply as $A$.

In addition to our analytical analysis, we have experimentally identified the
most computationally expensive kernels in GCN computation. As we will
demonstrate in Section~\ref{sec:runtime_decomposition}, we have profiled our single GPU
GCN training with nvprof to analyze the run-time of our kernels and pinpoint the
bottleneck kernels. We have observed that up to $94\%$ of the runtime
was spent during the execution of the forward and backward SpMM kernels.
Therefore, we have first focused on efficient parallelization of SpMM kernel on
multi-GPU setting.
Moving into multi-GPU from single GPU, one needs to find ways to distribute the data
into multiple GPUs, and adapt algorithms to perform parallel SpMM.

\subsection{Partitioning}

Given a matrix A, we can define 2D tiling (partitioning) of the matrix
using two partition vectors $p$ and $q$, such that $p$ represents the partition vector of
the first dimension, and $q$ represents the partition vector of the second
dimension. A partition vector $p$ with $P$ parts is defined as:
\begin{gather}
    p \in \mathbb{N}^{P + 1}, 0 = p(0) \leq \dots \leq p(i) \leq \dots \leq p(P) = n
\end{gather}
Then, let us define $A^{ij}$ as the $(i, j)$-th tile of the matrix.
\begin{gather}
    p(i) \leq u < p(i + 1), q(j) \leq v < q(j + 1) \\
    A^{ij}(u, v) = A(u + p(i), v + q(j)), u, v \in \mathbb{N}
\end{gather}

One way to partition $A$, $H$ and $W$ is to apply symmetric partitioning, so that $p = q$, to
the sparse matrix $A$, and then assign the tiles of $A$ to GPUs using 1D or
2D distribution. Let's start with 1D column distribution where $j$-th tile
column of $A$, $A^{ij}$, is assigned to $j$-th GPU.
Moreover, we will partition dense matrix $H$, using 1D partitioning by rows, with the same
partition vector $p$, and assign $H^{i}$ to $i$-th GPU.
Likewise, the resulting dense matrix will be conformally partitioned by its
rows. After partitioning, SpMM can be performed
in multiple stages. In each stage, one set of rows of the result matrix can
be filled, thus taking the algorithm $P$ steps to perform,
where $P$ is the number of GPUs. Each GPU performs an SpMM with their
local portions in the sparse and dense matrices.
That is, at stage $i$, $A^{ij}$ will be multiplied by $H^{j}$ by $j$-th GPU,
then partial results will be reduced at GPU $i$. \[
    C^i = \sum_j A^{ij}X^j
\]
The only communication needed for this operation is the reduction at the end.
In this scheme, $W$ is replicated across GPUs and is reduced at the end of every epoch of training.
The reduction of $W$ however is much faster than the communication done for the feature matrix $H$
because of their size difference $\mathcal{O}(d^2)$ vs $\mathcal{O}(nd)$.

\begin{figure}[h!]
    \centering
    \includegraphics[width=0.55\columnwidth]{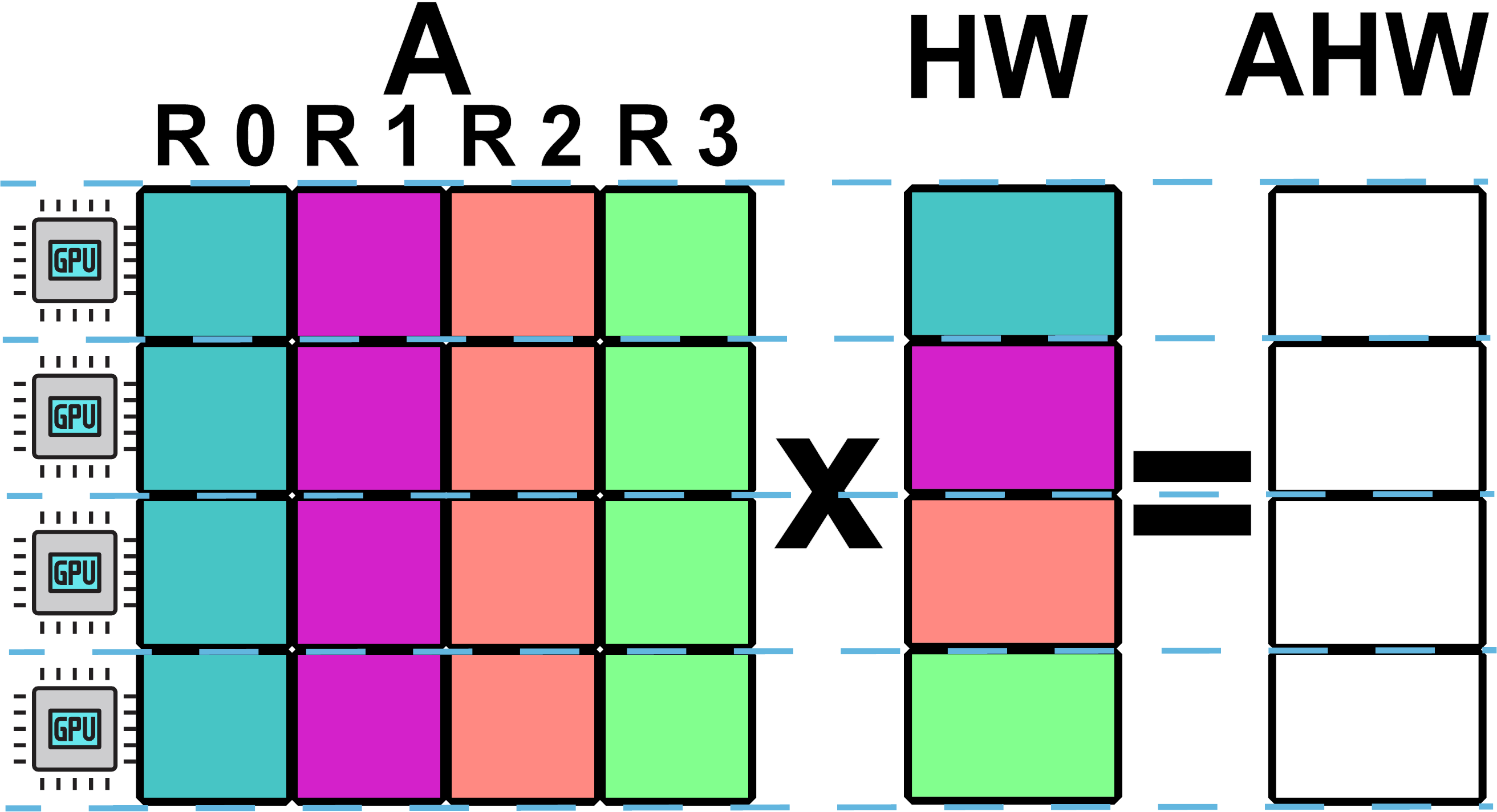}
    \caption{Partitioning of the sparse and dense matrices involved in SpMM. Colors represent stages, rows represent GPUs.}
    \label{fig:partitioning}
\end{figure}

\begin{figure}[h]
    \centering
    \begin{subfigure}[b]{0.475\columnwidth}
        \centering
        \includegraphics[width=\columnwidth]{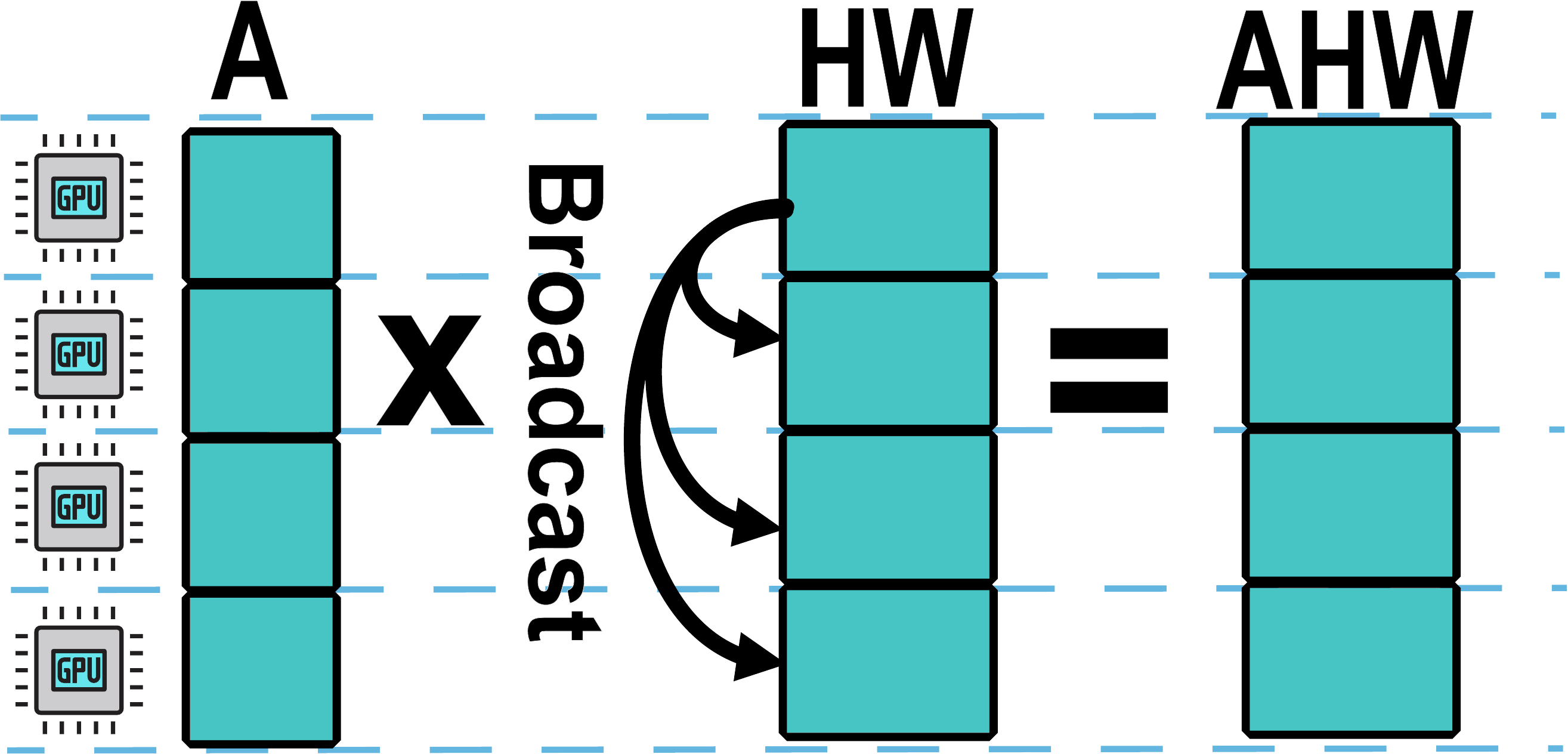}
        \caption{Example 1st stage}
        \label{fig:1stround}
    \end{subfigure}
    \vline
    \begin{subfigure}[b]{0.475\columnwidth}
        \centering
        \includegraphics[width=\columnwidth]{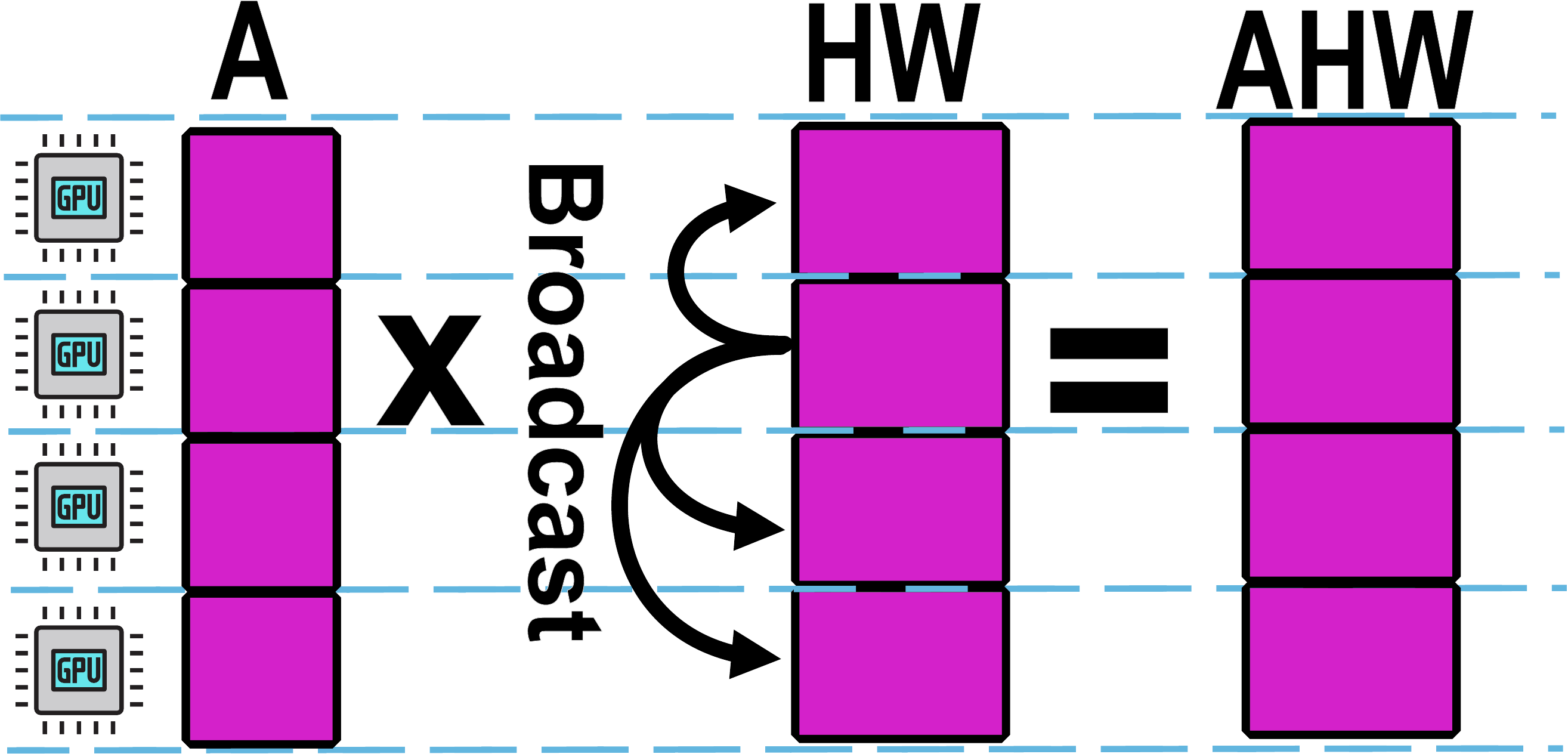}
        \caption{Example 2nd stage}
        \label{fig:2ndround}
    \end{subfigure}
    \caption{Example two stages of SpMM}
    \label{fig:example_rounds}
\end{figure}

Alternatively, one can do 1D row distribution and assigns $i$-th tile row of $A$, $A^{ij}$,
to $i$-th GPU, see Figure~\ref{fig:partitioning}.
Then, at stage $i$, $i$-th GPU broadcasts $H^{i}$, then $A^{ij}$ will
be multiplied by $H^{j}$ on $j$-th GPU. The only communication needed for this operation is
the broadcast at the beginning, see Figure~\ref{fig:example_rounds}.\[
    C^i = C^i + A^{ij}H^{j}
\]

Both of the above approaches partition $H$ by its rows, so one might consider
how it would work if $H$ was partitioned by its columns, into $1 \times P$ tiles.
For this case, let us use a partition vector $p$ with $P$ parts and a partition vector
$q$ with only a single part to partition $A$. Then, we can assign $A^{i1}$ to $i$-th GPU,
$H^{1j}$ to $j$-th GPU.
Likewise, this operation can be performed in multiple stages. At stage $i$,
$i$-th GPU broadcasts $A^{i1}$, then $A^{i1}$ will
be multiplied by $H^{1j}$ at $j$-th GPU. The results are kept at the $j$-th GPU.
The only communication needed for this
operation is the broadcast of the sparse matrix at the beginning.\[
    C^{ij} = A^{i1} H^{1j}
\]
However, for this particular partitioning strategy, there is more communication
involved during the GEMM kernel. In particular since $H$ is 1D column-partitioned,
$C^{ij} \times W^{jk}$ requires a reduction over $j$. This means
not only $A$ is communicated, but also the dense matrix $C$ is
communicated which makes this solution undesirable. Compared to the first solution, solution 2 provides better
load balance regardless of the matrix ordering, since each GPU is using the same
set of rows broadcasted at each stage, the sparsity pattern of the sparse
matrix will be identical across the GPUs. Nevertheless, since communication is the main bottleneck, we decide to
use the broadcast variant of solution 1, as in Figure~\ref{fig:partitioning}.

We don't discuss anymore complicated partitioning strategies such as 1.5D, 2D or 3D
as we will explain the reasoning in Section~\ref{subsec:choice_of_partition}. Furthermore, note
that the GeMM computations on the row-partitioned feature matrices do not require
any synchronization as each GPU can compute
$H^iW$ in~(\ref{eq:forward_pass_gemm}) independently.
The element-wise activation function is also fully independent, each GPU
computes it for their portions.

\subsection{Memory Optimizations}
\label{sec:memory_optim}

To reduce memory requirements, we reuse memory buffers in the forward
and backward passes, as much as possible.

\begin{figure}[h]

    \begin{subfigure}[b]{0.975\columnwidth}
        \centering
        \includegraphics[width=0.8\columnwidth]{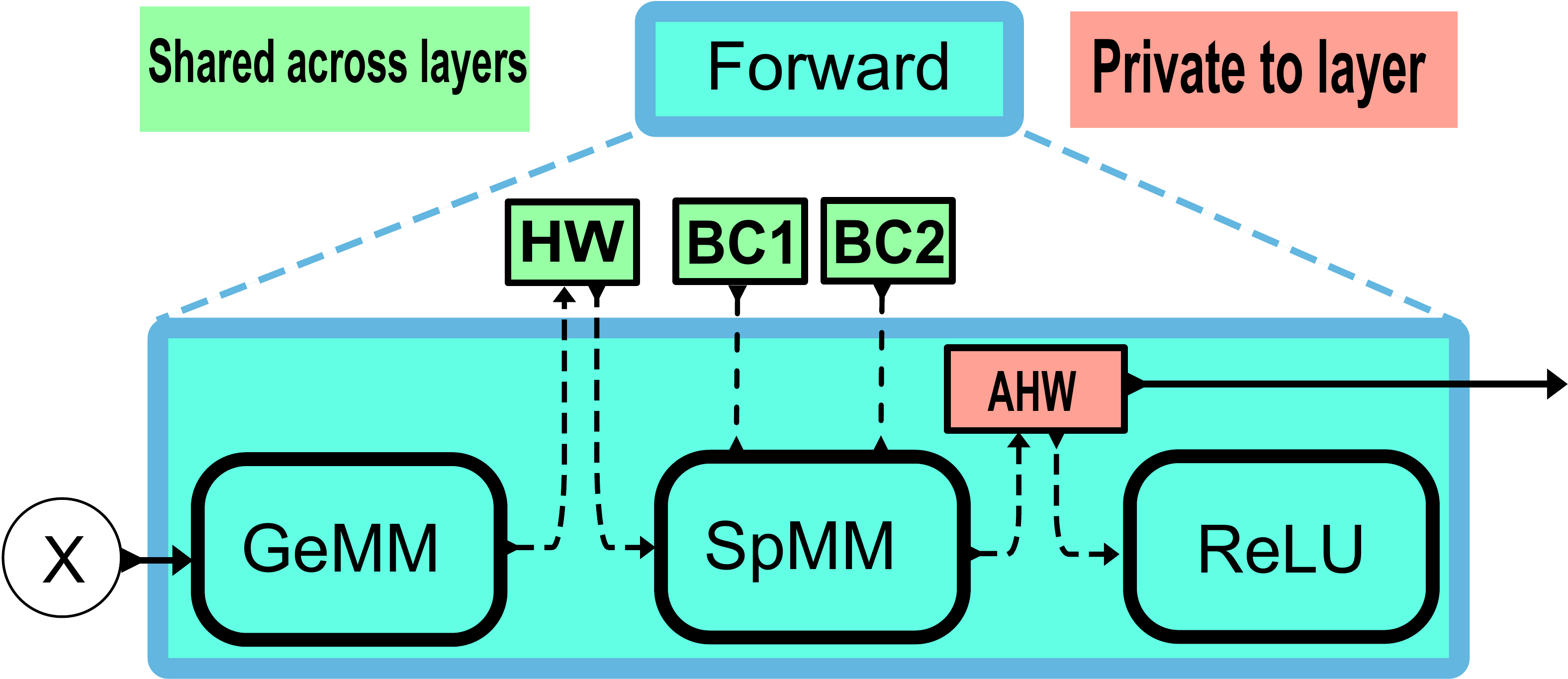}
        \caption{Forward Layer.}
        \label{fig:forward}
    \end{subfigure}
    \begin{subfigure}[b]{0.975\columnwidth}
        \centering
        \includegraphics[width=0.8\columnwidth]{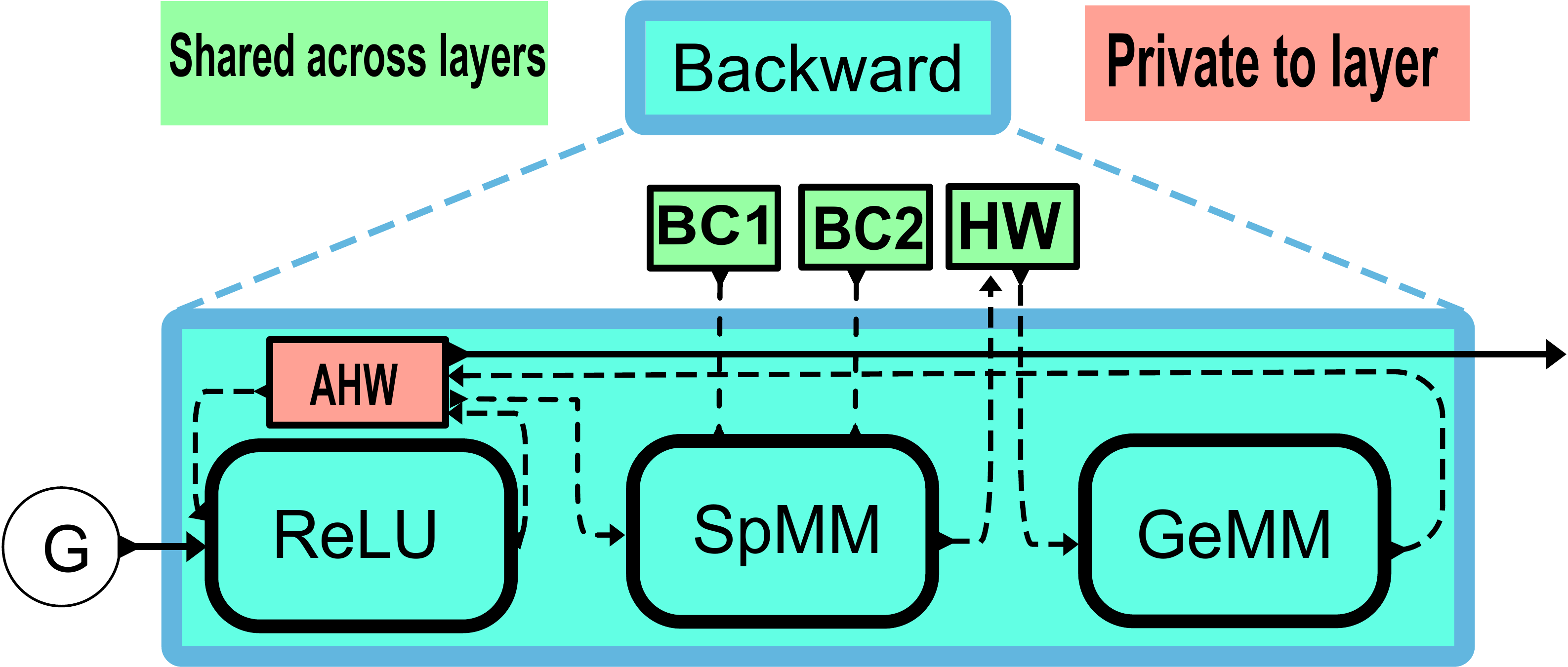}
        \caption{Backward Layer.}
        \label{fig:backward}
    \end{subfigure}
    \vspace*{-2ex}
    \caption{Forward and backward layers. Buffers are color according to being shared or private to layers. $BC1$, $BC2$, and $HW$
    are explained in Figure~\ref{fig:computation_diagram}. $AHW$: Buffer for
    result of the layer.}
    \label{fig:forward_backward}
\end{figure}

In the forward and backward computations
in~\cref{eq:forward_pass_gemm,eq:forward_pass_spmm,eq:forward_pass_activation},
we will have a temporary result buffer called $HW_B$ and a result buffer called
$AHW_B$ do the following mapping:
\begin{align}
    HW &\rightarrow HW_B \\
    AHW &\rightarrow AHW_B \\
    H' &\rightarrow AHW_B
\end{align}
And in the backward computations in~\cref{,eq:backward_pass_activation,eq:backward_pass_spmm,eq:backward_pass_gemm,eq:backward_pass_gemm1}:
\begin{align}
    AHW_G &\rightarrow AHW_B \\
    HW_G &\rightarrow HW_B \\
    H_G &\rightarrow AHW_B
\end{align}
Fig.~\ref{fig:forward} shows the mappings of the buffers
for the forward computation and Fig.~\ref{fig:backward} for the backward propagation.

Notice that, each layer only requires \emph{a single buffer} to store
their output. They also use a temporary buffer that
is shared across layers. Hence, each layer only increases the memory use by
a single buffer, compared to \emph{4x or 6x} in other
deep learning frameworks that allocates buffers for the output of SpMM,
GeMM and the activation functions. Considering the backward pass adds up-to
$6$ buffers per layer in total, as shown in Fig.~\ref{fig:computation_diagram}.
For $L$-layer GCN, the total number of buffers is $L + 3$, whose sizes on average are $n \times d$.

\subsection{Overlapping Computation and Communication}

Each round of our multi-round SpMM is composed of a broadcast of a tile of $H$
and a SpMM computation with a tile of $A$ with the received tile of $H$. Notice that,
there is an opportunity to overlap communication and computation in such a multi-round
scheme. After the broadcast of first tile of $H$, we overlap
communication of the next (remaining) tile(s) of $H$ with the SpMM computations.
In order to do that, we need an extra communication buffer for the {\em
next} $H$ tile. Since each GPU {\em keeps} its own $H$ tile, and
receives the $H^i$ in the $i$-th round, each GPU needs one more extra
buffer for the broadcast primitive. In total, overlapping communication and computation would
require two additional buffers.
In order to fully utilize communication computation
overlap, we use two GPU streams: one for communication (stream $1$) and one
computation (stream $0$). We launch all communication and computation kernels
asynchronously on those two streams and wait for $i$-th broadcast to finish on
stream $1$ before we start on $i$-th SpMM computation on stream $0$ and the $i+1$-th broadcast
waits for the $i-1$-th SpMM to finish not to overwrite its input when it is ongoing.

\subsection{Order of Computation and Saving one SpMM}

For computing $AHW$, we change the order of SpMM and GeMM operations depending
on the feature dimension of the current layer $d^{(l)}$ and the next layer
$d^{(l + 1)}$ as allowed by associativity. If $d^{(l)} < d^{(l + 1)}$, then
doing SpMM, otherwise running GeMM is faster.

If the gradients all the way back to the input features are not required, then
it is possible to skip the SpMM in the first layer during the backward pass. The
reason is that, SpMM scales each feature dimension independently so it is
possible to replace it with a diagonal feature scaling matrix in the first
layer's backward pass. In our case, each node takes the average of their
neighbors, thus the identity matrix is the scaling matrix, making it a
no-operation. Thus we avoid the SpMM of the first layer in the backward pass.

\section{Design Decisions}
\label{sec:design_decisions}

\subsection{Choice of the Partitioning Strategy}
\label{subsec:choice_of_partition}

The communication topology of the system directly effects observed bandwidth of
different communication patterns. This is clearly not an issue for systems like
DGX-A100 where 8 GPU of the system are connected shared NVlink switch with 12
links, and could achieve full communication bandwidth between any pair of GPUs.
Whereas, in DGX-1 there are only 6 links and connections between GPUs are
asymmetric. Such asymmetry will make some theoretically optimum algorithms
perform poorly on that system, since the underlying communication assumptions
are not valid on that system. For example, 1.5D algorithm presented
in~\cite{CAGNET} halves the theoretical communication volume, by using more
memory with replication factor $c = 2$.
If we group the GPUs into two groups as per the
replication factor, each group has 4 links available. Then the broadcast can be
faster by a factor of $\frac{6}{4 \times 2}$ in the 1.5D case. However, the
last reduction among the two groups has access to only 2 links. Then, if we sum
up the time required for communication for the 1.5D case, which necessitates two
rounds of broadcast followed by a concurrent reduction (see ~\cite{CAGNET} for
the details of the algorithm) we get:
$2\frac{nd}{4 \times 4l} + \frac{nd}{4 \times 2l} = \frac{nd}{4l}$, where $l$ is
the single NVlink bandwidth.
In comparison, the 1D algorithm only takes $8\frac{nd}{8 \times 6l} =
\frac{nd}{6l}$ time. On the other hand, 
in DGX-A100 all broadcasts and reductions can utilize all of the 12
links. Hence, summing up time required for the 1.5D algorithm, we get:
$2\frac{nd}{4 \times 12l} + \frac{nd}{4 \times 12l} = \frac{nd}{16l}$.
In comparison, the 1D algorithm takes $8\frac{nd}{8 \times 12l} =
\frac{nd}{12l}$ time.

According to the above analysis, 1.5D algorithm is slower on DGX-1 by a factor
of $\frac{2}{3}$ but it is faster on DGX A100 by $\frac{4}{3}$, but also
requires twice as much memory. Since GNN training is usually bound by the
GPU memory, we chose to implement only the 1D version.

\subsection{Permutation}

In order to balance the number of nonzeros in each part $A^{ij}$ in the
uniformly partitioned sparse matrices, we randomly permute their vertices.
This has a significant effect on load balance compared to using the original
orderings of the sparse matrices which can have highly imbalanced parts.
Later in Section~\ref{subsec:permutation}, 
we show how this permutation improves the execution time with better load
balancing, especially with larger number of GPUs.

\section{Experimental Evaluation}
\label{sec:experiments}

\subsection{Experiment Setup}
\label{sec:expr:setup}

\noindent{\em Hardware and Software:}
We perform our experiments on two machines: NVIDIA DGX-1, also referred to as DGX-V100, and NVIDIA DGX-A100.
DGX-V100 has $8$ Nvidia V100 GPUs, each equipped with $32$GB memory with a
$900$ GB/s memory bandwidth. Each V100 has $6$ NVLink connections, each
consisting of $2$ sub-links that send data in one direction, and has a
$25$GB/s bandwidth. That is, each link is capable of $50$GB/s bidirectional
bandwidth, and theoretically, the aggregate system bandwidth is $300$ GB/s.
The DGX-1 is equipped with a dual 20-core Intel Xeon E5-2698 CPU with
$512$ GB RAM. NVIDIA DGX-A100 has $8$ NVIDIA A100 GPUs, each equipped with $80$GB memory with a $2$
TB/s memory bandwidth. Each A100 has $12$ NVLink connections,
thus twice as much the bandwidth of V100. Unlike the V100, each A100 is connected to an
NWSwitch, enabling a full peer-to-peer bidirectional bandwidth of $600$ GB/s
between any two GPUs. DGX-A100 is equipped with a dual 64-core AMD Rome 7742
CPU with $2$ TB RAM. Both machines run Ubuntu $20.04$.

We implemented MG-GCN using C++ standard 20 and compiled with GCC 9.3.0
and CUDA 11.4. We used CUDA's cuSPARSE for SpMM calls with
the Compressed Sparse Row format for the sparse matrices, and cuBLAS
for GeMM with the Row Major format for the dense matrices. PIGO~\cite{Gabert21-GrAPL} is
used for IO purposes.
We use DGL 0.7.1 which is currently the latest available version~\cite{wang2019dgl}.
We follow the instructions for compiling CAGNET~\cite{CAGNET} on its repository.
For MG-GCN, we use NCCL
(Nvidia Collective Communication Library) 2.11.4
and for CAGNET, we use NCCL 2.4.8 for compatibility reasons.
The code for MG-GCN is available at \url{https://github.com/GT-TDAlab/MG-GCN}.

\noindent{\em Datasets:}
We use two types of datasets in our experiment. The first category is
GNN Benchmark datasets which are popular datasets used in GNN research,
see Table~\ref{tab:Datasets}.
The Reddit dataset is a graph from Reddit posts that are posted in September,
2014~\cite{hamilton2018inductive}. The node labels represent the communities (subreddits).
Products (OGBN-Products) is a graph from Amazon co-purchase network.
Nodes represents the products, and link represents products
that are bought together. Proteins (OGBN-Proteins) is a biological network graph
dataset where nodes represents proteins and edges represents associations
between proteins. Arxiv (OGBN-Arxiv) and Cora are citation networks where each
node represent a paper and directed edges represent citation
direction~\cite{sen:aimag08, hu2021open}.

\begin{table}[h]
    \caption{\textbf{Benchmark Datasets}. $n$: \#vertices, $m$: \#edges, $d^{(0)}$: \#features, $d^{(L)}$: \#classes, $k$: average degree.}
    \label{tab:Datasets}
    \vskip 0.15in
    \begin{center}
    \begin{small}
    \begin{sc}
    \begin{tabular}{crrrrr}
    \toprule
    Dataset & $n$ & $m$ & $d^{(0)}$ & $d^{(L)}$ & $k$\\
    \midrule
     Cora & 3.3K & 9.2K  & 3.7K & 6 & 3\\
     Arxiv & 169K & 1.16M  & 128 & 40 & 7\\
     Papers & 111M  &  1.61B  & 128 & 172& 15\\
     Products & 2.5M  & 126M & 104 & 47 & 52\\
     Proteins & 8.74M  & 1.3B  & 128 & 256& 150\\
     Reddit & 233K & 115M & 602 & 41 & 492\\
    \bottomrule
    \end{tabular}
    \end{sc}
    \end{small}
    \end{center}
    \vskip -0.1in
\end{table}

We also used synthetic datasets generated with
BTER~\cite{Kolda_2014} to evaluate scalability of our method under varying density.
BTER requires a degree distribution and clustering
coefficient by degree as input and generates synthetic graphs matching those
properties. We first profile the degree distribution of Arxiv dataset,
then by increasing the average degree and fixing the number of vertices,
we generate $8$ synthetic datasets. We name these datasets as $1x, \ldots, 128x$.
As the name suggests, the number represents the
scaling factor of number of edges from the original graph.
We generate the features and assign class labels randomly.
Each synthetic dataset has a feature vector of size
$512$, and there are $40$ classes.
Since the graphs generated by BTER are not deterministic,
we generate $10$ of each scale, and take the median while reporting the results.

\label{sec:Model}
\noindent{\em Model:}
While we are able to train more complex models, to make fair comparisons,
we use $4$ different GCN models. First, to compare with
CAGNET and DGL, we use a model with $2$ layers, and the hidden layer consists
of $512$ neurons. Our limitation comes from the fact that, the available code
for CAGNET does not have the option to change the number of layers. Second, to
compare with DistGNN  on Reddit, we use a model with $2$ layers and hidden
layer consists $16$ neurons. To compare with DistGNN on Products, Protein and
Papers, we use a model with $3$ layers and hidden layers consist of $256$ neurons.
Finally, we use a 4th model with $3$ layer, each consisting of $208$ neurons
to run MG-GCN on Papers DGX-A100, since $208$ is
the largest hidden layer size that can fit into DGX-A100. We have implemented and
used the Adam optimizer~\cite{kingma2017adam} and the softmax cross entropy
loss~\cite{logistic} in all of our experiments.

\subsection{Runtime Breakdown of GCN Computation}
\label{sec:runtime_decomposition}

\begin{figure}[ht!]
    \centering
    \includegraphics[width=0.98\columnwidth]{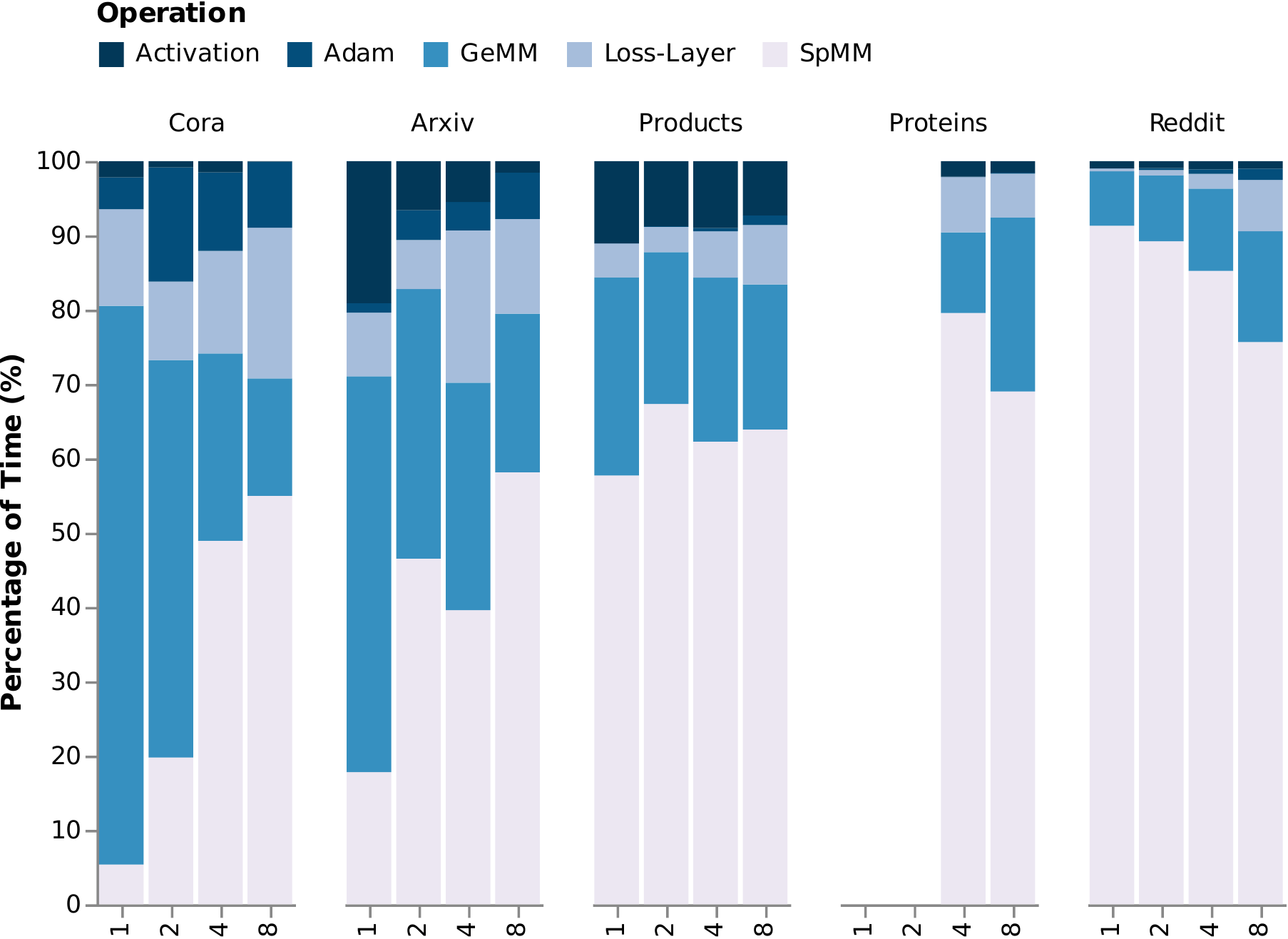}
    \caption{Runtime decomposition of operations involved in forward and backward pass.}
    \label{fig:runtime_decomposition}
\end{figure}

We analyze the breakdown of execution time of GCN computation in order to
find the computational bottlenecks during training.
Figure~\ref{fig:runtime_decomposition} presents the runtime breakdown of the first
GCN model described in Section~\ref{sec:Model}. The activation layer refers to the
computation in~\cref{eq:forward_pass_activation}, Adam refers to the update of the
model parameters $W$ by the Adam optimizer and loss layer refers to the computations
related to the softmax cross entropy loss. As it is evident from the
figure, for sufficiently large datasets, i.e., Proteins, Products, and Reddit,
the main bottleneck is SpMM kernel which takes 60\%-94\% of the runtime,
and second bottleneck is GeMM kernel 5\%-20\% of the runtime. On the other hand,
for small datasets the main bottleneck becomes GeMM. Therefore, we stress the
importance of parallelizing SpMM and GeMM kernels to achieve scalability during
any GCN training, and focus our attention to parallelizing these kernels.

\subsection{Impact of Permutation}
\label{subsec:permutation}

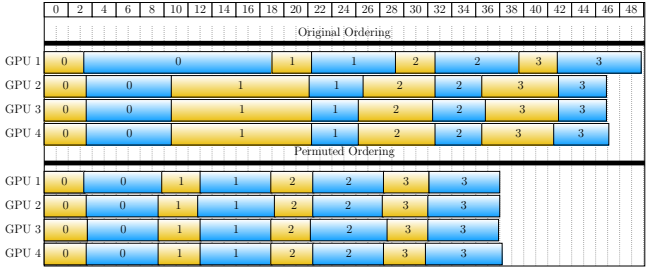
\begin{figure}[h]
    \centering
    \begin{minipage}{\columnwidth}
    \resizebox{\columnwidth}{!}{
    \begin{tikzpicture}
        \begin{ganttchart}[       
            x unit=0.05cm,
            vgrid={*9{draw=none}, dotted},
            y unit chart=1.0cm,
            inline
        ]{0}{500}
            \gantttitle[title label font=\Large]{0}{20}
            \gantttitle[title label font=\Large]{2}{20}
            \gantttitle[title label font=\Large]{4}{20}
            \gantttitle[title label font=\Large]{6}{20}
            \gantttitle[title label font=\Large]{8}{20}
            \gantttitle[title label font=\Large]{10}{20}
            \gantttitle[title label font=\Large]{12}{20}
            \gantttitle[title label font=\Large]{14}{20}
            \gantttitle[title label font=\Large]{16}{20}
            \gantttitle[title label font=\Large]{18}{20}
            \gantttitle[title label font=\Large]{20}{20}
            \gantttitle[title label font=\Large]{22}{20}
            \gantttitle[title label font=\Large]{24}{20}
            \gantttitle[title label font=\Large]{26}{20}
            \gantttitle[title label font=\Large]{28}{20}
            \gantttitle[title label font=\Large]{30}{20}
            \gantttitle[title label font=\Large]{32}{20}
            \gantttitle[title label font=\Large]{34}{20}
            \gantttitle[title label font=\Large]{36}{20}
            \gantttitle[title label font=\Large]{38}{20}
            \gantttitle[title label font=\Large]{40}{20}
            \gantttitle[title label font=\Large]{42}{20}
            \gantttitle[title label font=\Large]{44}{20}
            \gantttitle[title label font=\Large]{46}{20}
            \gantttitle[title label font=\Large]{48}{20}            
            \ganttnewline
            \ganttgroup[group label font=\Large, group top shift=.6]{Original Ordering}{0}{500}
            \ganttnewline
            \ganttcompbar[inline=false]{GPU 1}{0.3999999999996362}{33.69999999999891}
            \ganttcommbar[]{0}{0.3999999999996362}{33.69999999999891}
            \ganttcompbar[]{0}{33.69999999999891}{190.0999999999999}
            \ganttcommbar[]{1}{190.19999999999982}{223.599999999999}
            \ganttcompbar[]{1}{223.599999999999}{293.6999999999989}
            \ganttcommbar[]{2}{293.6999999999989}{326.80000000000064}
            \ganttcompbar[]{2}{326.80000000000064}{396.7000000000007}
            \ganttcommbar[]{3}{396.80000000000064}{428.4999999999991}
            \ganttcompbar[]{3}{428.4999999999991}{497.89999999999964}
            \ganttnewline
            \ganttcompbar[inline=false]{GPU 2}{0.09999999999990905}{35.99999999999909}
            \ganttcommbar[]{0}{0.09999999999990905}{35.99999999999909}
            \ganttcompbar[]{0}{35.99999999999909}{106.60000000000082}
            \ganttcommbar[]{1}{106.60000000000082}{221.29999999999882}
            \ganttcompbar[]{1}{221.29999999999882}{266.099999999999}
            \ganttcommbar[]{2}{266.1999999999989}{326.90000000000055}
            \ganttcompbar[]{2}{326.90000000000055}{365.8999999999992}
            \ganttcommbar[]{3}{365.8999999999992}{429.40000000000055}
            \ganttcompbar[]{3}{429.40000000000055}{468.4999999999991}
            \ganttnewline
            \ganttcompbar[inline=false]{GPU 3}{0.09999999999990905}{35.99999999999909}
            \ganttcommbar[]{0}{0.09999999999990905}{35.99999999999909}
            \ganttcompbar[]{0}{35.99999999999909}{106.29999999999882}
            \ganttcommbar[]{1}{106.39999999999873}{223.6999999999989}
            \ganttcompbar[]{1}{223.6999999999989}{262.89999999999964}
            \ganttcommbar[]{2}{262.89999999999964}{324.30000000000064}
            \ganttcompbar[]{2}{324.40000000000055}{368.99999999999864}
            \ganttcommbar[]{3}{368.99999999999864}{429.40000000000055}
            \ganttcompbar[]{3}{429.40000000000055}{468.4999999999991}
            \ganttnewline
            \ganttcompbar[inline=false]{GPU 4}{0.0}{35.79999999999927}
            \ganttcommbar[]{0}{0.0}{35.79999999999927}
            \ganttcompbar[]{0}{35.79999999999927}{106.29999999999882}
            \ganttcommbar[]{1}{106.29999999999882}{223.4999999999991}
            \ganttcompbar[]{1}{223.4999999999991}{262.5999999999999}
            \ganttcommbar[]{2}{262.5999999999999}{326.6000000000008}
            \ganttcompbar[]{2}{326.6000000000008}{365.59999999999945}
            \ganttcommbar[]{3}{365.69999999999936}{425.9999999999991}
            \ganttcompbar[]{3}{425.9999999999991}{470.69999999999936}
            \ganttnewline
            \ganttgroup[group label font=\Large, group top shift=.6]{Permuted Ordering}{0}{500}
            \ganttnewline
            \ganttcompbar[inline=false]{GPU 1}{0.1999999999998181}{33.400000000001455}
            \ganttcommbar[]{0}{0.1999999999998181}{33.400000000001455}
            \ganttcompbar[]{0}{33.500000000001364}{98.09999999999945}
            \ganttcommbar[]{1}{98.20000000000164}{130.70000000000164}
            \ganttcompbar[]{1}{130.70000000000164}{189.40000000000055}
            \ganttcommbar[]{2}{189.40000000000055}{224.70000000000027}
            \ganttcompbar[]{2}{224.70000000000027}{283.30000000000155}
            \ganttcommbar[]{3}{283.30000000000155}{320.90000000000146}
            \ganttcompbar[]{3}{321.00000000000136}{379.60000000000036}
            \ganttnewline
            \ganttcompbar[inline=false]{GPU 2}{0.09999999999990905}{35.900000000001455}
            \ganttcommbar[]{0}{0.09999999999990905}{35.900000000001455}
            \ganttcompbar[]{0}{35.900000000001455}{95.39999999999964}
            \ganttcommbar[]{1}{95.49999999999955}{128.40000000000146}
            \ganttcompbar[]{1}{128.40000000000146}{192.4000000000001}
            \ganttcommbar[]{2}{192.4000000000001}{224.20000000000073}
            \ganttcompbar[]{2}{224.30000000000064}{282.6999999999998}
            \ganttcommbar[]{3}{282.6999999999998}{320.59999999999945}
            \ganttcompbar[]{3}{320.70000000000164}{379.1000000000008}
            \ganttnewline
            \ganttcompbar[inline=false]{GPU 3}{0.0}{35.900000000001455}
            \ganttcommbar[]{0}{0.0}{35.900000000001455}
            \ganttcompbar[]{0}{36.000000000001364}{95.39999999999964}
            \ganttcommbar[]{1}{95.39999999999964}{130.90000000000146}
            \ganttcompbar[]{1}{130.90000000000146}{189.40000000000055}
            \ganttcommbar[]{2}{189.40000000000055}{222.00000000000045}
            \ganttcompbar[]{2}{222.00000000000045}{286.00000000000136}
            \ganttcommbar[]{3}{286.00000000000136}{320.2999999999997}
            \ganttcompbar[]{3}{320.39999999999964}{378.8000000000011}
            \ganttnewline
            \ganttcompbar[inline=false]{GPU 4}{0.0}{35.70000000000164}
            \ganttcommbar[]{0}{0.0}{35.70000000000164}
            \ganttcompbar[]{0}{35.70000000000164}{95.09999999999991}
            \ganttcommbar[]{1}{95.09999999999991}{130.70000000000164}
            \ganttcompbar[]{1}{130.70000000000164}{189.20000000000073}
            \ganttcommbar[]{2}{189.20000000000073}{224.50000000000045}
            \ganttcompbar[]{2}{224.50000000000045}{283.09999999999945}
            \ganttcommbar[]{3}{283.09999999999945}{318.09999999999945}
            \ganttcompbar[]{3}{318.09999999999945}{381.80000000000064}
        \end{ganttchart}
        \end{tikzpicture}
    }
    \end{minipage}
    \caption{Timeline of the SpMM on the Products dataset using its original and
    permuted ordering. The numbers on the bars represent stages. For each GPU, computation (blue) and communication (yellow)
    phases are separately plotted.}
    \label{fig:original_vs_permuted_timeline}
\end{figure}

\begin{figure}[h]
    \centering
    \includegraphics[width=0.98\columnwidth]{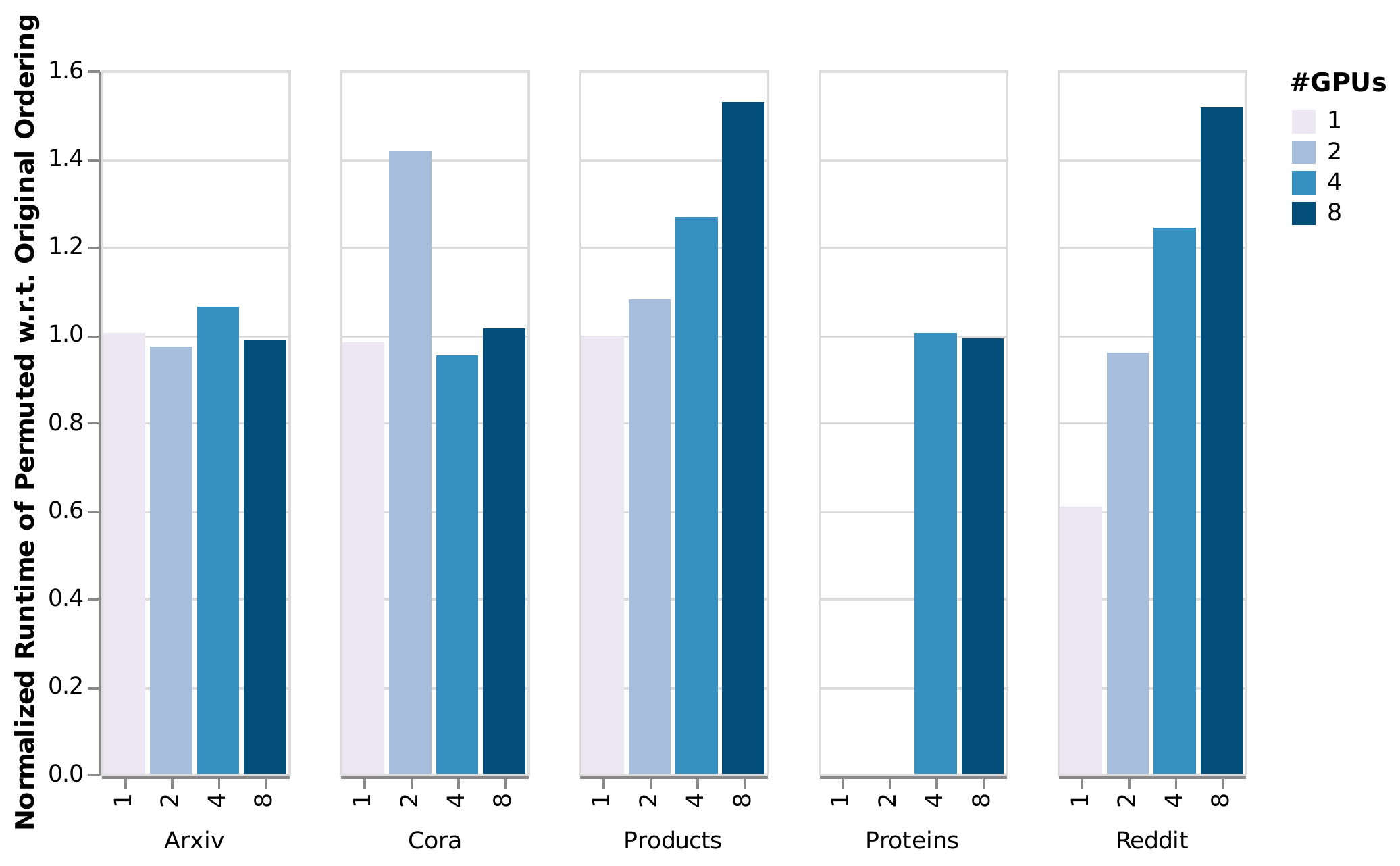}
    \caption{Effect of permuting to epoch runtime on DGX-V100.}
    \label{fig:permute_results}
\end{figure}

Figure~\ref{fig:original_vs_permuted_timeline} presents the breakdown of execution of SpMM
to communication and computation times for each stage for the Product dataset
using original and permuted ordering. On the top part of figure, there is a significant
computational imbalance that hampers the efficient parallel execution. To
remedy load imbalance problem we randomly permute the adjacency matrix
before the computation. On the bottom part of the figure, permuted ordering
achieves better computation load balance and reduces the execution time from
50ms to 38ms. Figure~\ref{fig:permute_results} shows normalized runtime improvement of
permuted ordering w.r.t. original ordering for each dataset for varying number of GPUs.
As seen in the figure, permutation yields slightly slower
execution time on small number of GPUs for some dataset;
however, as number of GPUs increases, the runtime improves significantly
with the load balance achieved by permutation. For example,
we observed $1.5 \times$ runtime improvement on Products
and Reddit datasets with $8$ GPUs.

\definecolor{commcolor}{RGB}{234, 187, 0}

\definecolor{compcolor}{RGB}{0, 153, 255}

\newganttchartelement{commbar}{
    commbar/.style={
        shape=rectangle,
        inner sep=0pt,
        draw=black,
       top color=white,
       bottom color=commcolor!95,
    },
    commbar inline label anchor=center,
    commbar inline label node/.append style={anchor=center, text=black},
    commbar height=.9,
    commbar top shift = 0.05,
    commbar label font=\Large
}

\newganttchartelement{compbar}{
    compbar/.style={
        shape=rectangle,
        inner sep=0pt,
        draw=black,
      top color=white, 
      bottom color=compcolor!95,
    },
    compbar inline label anchor=center,
    compbar inline label node/.append style={anchor=center, text=black},
    compbar height=.9,
    compbar top shift = 0.05,
    compbar label font=\Large
}

\subsection{Overlapping Computation and Communication}

\begin{figure}
    \centering
    \begin{minipage}{\columnwidth}
    \resizebox{\columnwidth}{!}{
        \begin{tikzpicture}
            \begin{ganttchart}[       
                x unit=0.05cm,
                vgrid={*9{draw=none}, dotted},
                y unit chart=1.0cm,
                inline
            ]{0}{400}
                \gantttitle[title label font=\Large]{0}{20}
                \gantttitle[title label font=\Large]{2}{20}
                \gantttitle[title label font=\Large]{4}{20}
                \gantttitle[title label font=\Large]{6}{20}
                \gantttitle[title label font=\Large]{8}{20}
                \gantttitle[title label font=\Large]{10}{20}
                \gantttitle[title label font=\Large]{12}{20}
                \gantttitle[title label font=\Large]{14}{20}
                \gantttitle[title label font=\Large]{16}{20}
                \gantttitle[title label font=\Large]{18}{20}
                \gantttitle[title label font=\Large]{20}{20}
                \gantttitle[title label font=\Large]{22}{20}
                \gantttitle[title label font=\Large]{24}{20}
                \gantttitle[title label font=\Large]{26}{20}
                \gantttitle[title label font=\Large]{28}{20}
                \gantttitle[title label font=\Large]{30}{20}
                \gantttitle[title label font=\Large]{32}{20}
                \gantttitle[title label font=\Large]{34}{20}
                \gantttitle[title label font=\Large]{36}{20}
                \gantttitle[title label font=\Large]{38}{20}
                \ganttnewline
                \ganttcompbar[inline=false]{GPU 1 (no over)}{0.1999999999998181}{33.400000000001455}
                \ganttcommbar[]{0}{0.1999999999998181}{33.400000000001455}
                \ganttcompbar[]{0}{33.500000000001364}{98.09999999999945}
                \ganttcommbar[]{1}{98.20000000000164}{130.70000000000164}
                \ganttcompbar[]{1}{130.70000000000164}{189.40000000000055}
                \ganttcommbar[]{2}{189.40000000000055}{224.70000000000027}
                \ganttcompbar[]{2}{224.70000000000027}{283.30000000000155}
                \ganttcommbar[]{3}{283.30000000000155}{320.90000000000146}
                \ganttcompbar[]{3}{321.00000000000136}{379.60000000000036}
                \ganttnewline
                \ganttcompbar[inline=false]{comm 1}{0.29999999999972715}{33.70000000000118}
                \ganttcommbar[]{0}{0.29999999999972715}{33.70000000000118}
                \ganttcommbar[]{1}{33.80000000000109}{79.20000000000073}
                \ganttcommbar[]{2}{108.900000000001}{152.5999999999999}
                \ganttcommbar[]{3}{178.30000000000155}{223.20000000000164}
                \ganttnewline
                \ganttcompbar[inline=false]{comp 1}{33.900000000001}{108.80000000000109}
                \ganttcompbar[]{0}{33.900000000001}{108.80000000000109}
                \ganttcompbar[]{1}{108.80000000000109}{178.20000000000164}
                \ganttcompbar[]{2}{178.20000000000164}{247.5}
                \ganttcompbar[]{3}{247.5}{306.3000000000011}
                \ganttnewline
                \ganttnewline
                \ganttcompbar[inline=false]{GPU 2 (no over)}{0.09999999999990905}{35.900000000001455}
                \ganttcommbar[]{0}{0.09999999999990905}{35.900000000001455}
                \ganttcompbar[]{0}{35.900000000001455}{95.39999999999964}
                \ganttcommbar[]{1}{95.49999999999955}{128.40000000000146}
                \ganttcompbar[]{1}{128.40000000000146}{192.4000000000001}
                \ganttcommbar[]{2}{192.4000000000001}{224.20000000000073}
                \ganttcompbar[]{2}{224.30000000000064}{282.6999999999998}
                \ganttcommbar[]{3}{282.6999999999998}{320.59999999999945}
                \ganttcompbar[]{3}{320.70000000000164}{379.1000000000008}
                \ganttnewline
                \ganttcompbar[inline=false]{comm 2}{0.09999999999990905}{35.800000000001546}
                \ganttcommbar[]{0}{0.09999999999990905}{35.800000000001546}
                \ganttcommbar[]{1}{35.800000000001546}{74.30000000000064}
                \ganttcommbar[]{2}{104.20000000000073}{152.5999999999999}
                \ganttcommbar[]{3}{178.8000000000011}{222.5999999999999}
                \ganttnewline
                \ganttcompbar[inline=false]{comp 2}{35.900000000001455}{104.10000000000082}
                \ganttcompbar[]{0}{35.900000000001455}{104.10000000000082}
                \ganttcompbar[]{1}{104.10000000000082}{178.60000000000127}
                \ganttcompbar[]{2}{178.70000000000118}{247.5999999999999}
                \ganttcompbar[]{3}{247.5999999999999}{306.1000000000013}
                \ganttnewline
                \ganttnewline
                \ganttcompbar[inline=false]{GPU 3 (no over)}{0.0}{35.900000000001455}
                \ganttcommbar[]{0}{0.0}{35.900000000001455}
                \ganttcompbar[]{0}{36.000000000001364}{95.39999999999964}
                \ganttcommbar[]{1}{95.39999999999964}{130.90000000000146}
                \ganttcompbar[]{1}{130.90000000000146}{189.40000000000055}
                \ganttcommbar[]{2}{189.40000000000055}{222.00000000000045}
                \ganttcompbar[]{2}{222.00000000000045}{286.00000000000136}
                \ganttcommbar[]{3}{286.00000000000136}{320.2999999999997}
                \ganttcompbar[]{3}{320.39999999999964}{378.8000000000011}
                \ganttnewline
                \ganttcompbar[inline=false]{comm 3}{0.0}{35.800000000001546}
                \ganttcommbar[]{0}{0.0}{35.800000000001546}
                \ganttcommbar[]{1}{35.900000000001455}{78.60000000000127}
                \ganttcommbar[]{2}{106.60000000000082}{147.99999999999955}
                \ganttcommbar[]{3}{173.20000000000164}{222.5}
                \ganttnewline
                \ganttcompbar[inline=false]{comp 3}{35.900000000001455}{106.50000000000091}
                \ganttcompbar[]{0}{35.900000000001455}{106.50000000000091}
                \ganttcompbar[]{1}{106.50000000000091}{173.10000000000173}
                \ganttcompbar[]{2}{173.10000000000173}{247.5}
                \ganttcompbar[]{3}{247.5}{306.2000000000012}
                \ganttnewline
                \ganttnewline
                \ganttcompbar[inline=false]{GPU 4 (no over)}{0.0}{35.70000000000164}
                \ganttcommbar[]{0}{0.0}{35.70000000000164}
                \ganttcompbar[]{0}{35.70000000000164}{95.09999999999991}
                \ganttcommbar[]{1}{95.09999999999991}{130.70000000000164}
                \ganttcompbar[]{1}{130.70000000000164}{189.20000000000073}
                \ganttcommbar[]{2}{189.20000000000073}{224.50000000000045}
                \ganttcompbar[]{2}{224.50000000000045}{283.09999999999945}
                \ganttcommbar[]{3}{283.09999999999945}{318.09999999999945}
                \ganttcompbar[]{3}{318.09999999999945}{381.80000000000064}
                \ganttnewline
                \ganttcompbar[inline=false]{comm 4}{0.0}{35.800000000001546}
                \ganttcommbar[]{0}{0.0}{35.800000000001546}
                \ganttcommbar[]{1}{35.800000000001546}{78.60000000000127}
                \ganttcommbar[]{2}{106.20000000000118}{152.4000000000001}
                \ganttcommbar[]{3}{175.29999999999973}{216.5000000000009}
                \ganttnewline
                \ganttcompbar[inline=false]{comp 4}{35.900000000001455}{106.00000000000136}
                \ganttcompbar[]{0}{35.900000000001455}{106.00000000000136}
                \ganttcompbar[]{1}{106.10000000000127}{175.19999999999982}
                \ganttcompbar[]{2}{175.19999999999982}{241.90000000000055}
                \ganttcompbar[]{3}{241.90000000000055}{305.6000000000017}
            \end{ganttchart}
        \end{tikzpicture}
    }
    \end{minipage}
    \caption{Timeline of the SpMM on the Products dataset using permuted ordering. The numbers on the bars represent stages. Each GPU is represented by 3 lines. First line represents computation without overlapping communication. Next two lines represent computation with overlapping communication. Blue line: computation time. Yellow line: communication time.}
    \label{fig:combined_timeline}
\end{figure}
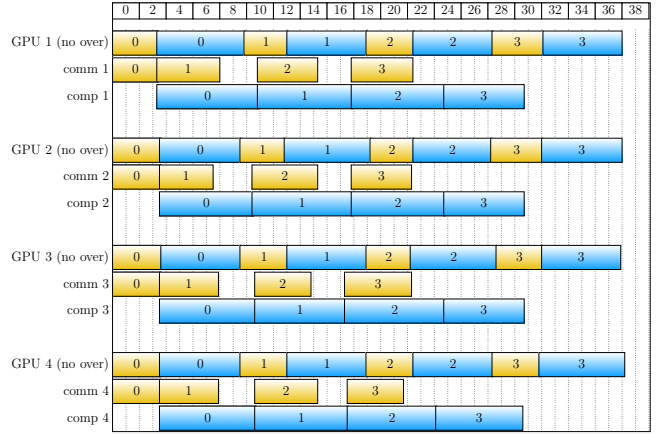

Figure~\ref{fig:combined_timeline} shows the effect of the
communication-computation overlap on Products datasets using 4 GPUs.
Notice that overlapping these two operations makes both the computation
and the communication slower. This is because of the use of shared
resources, in particular the memory bandwidth. Since SpMM is a mostly memory
bandwidth bound operation, it becomes slower when overlapped with the
communication kernel that takes up some of the
global memory bandwidth. The global memory bandwidth of a V100 GPU is
$900$ GB/s and the communication bandwidth is
$150$ GB/s. Assuming the communication happens at full bandwidth,
this results into a reduction of the global memory
bandwidth for the SpMM operation by a factor of $\frac{1}{6}$.
Nevertheless, communication-computation overlap still improves the performance.
As seen in the figure, for Products, SpMM time can be reduced to 30ms from 38ms
with overlapping communication and computation.

\subsection{Impact of Average Degree}

\begin{figure}[h]
    \centering
    \includegraphics[width=0.50\columnwidth]{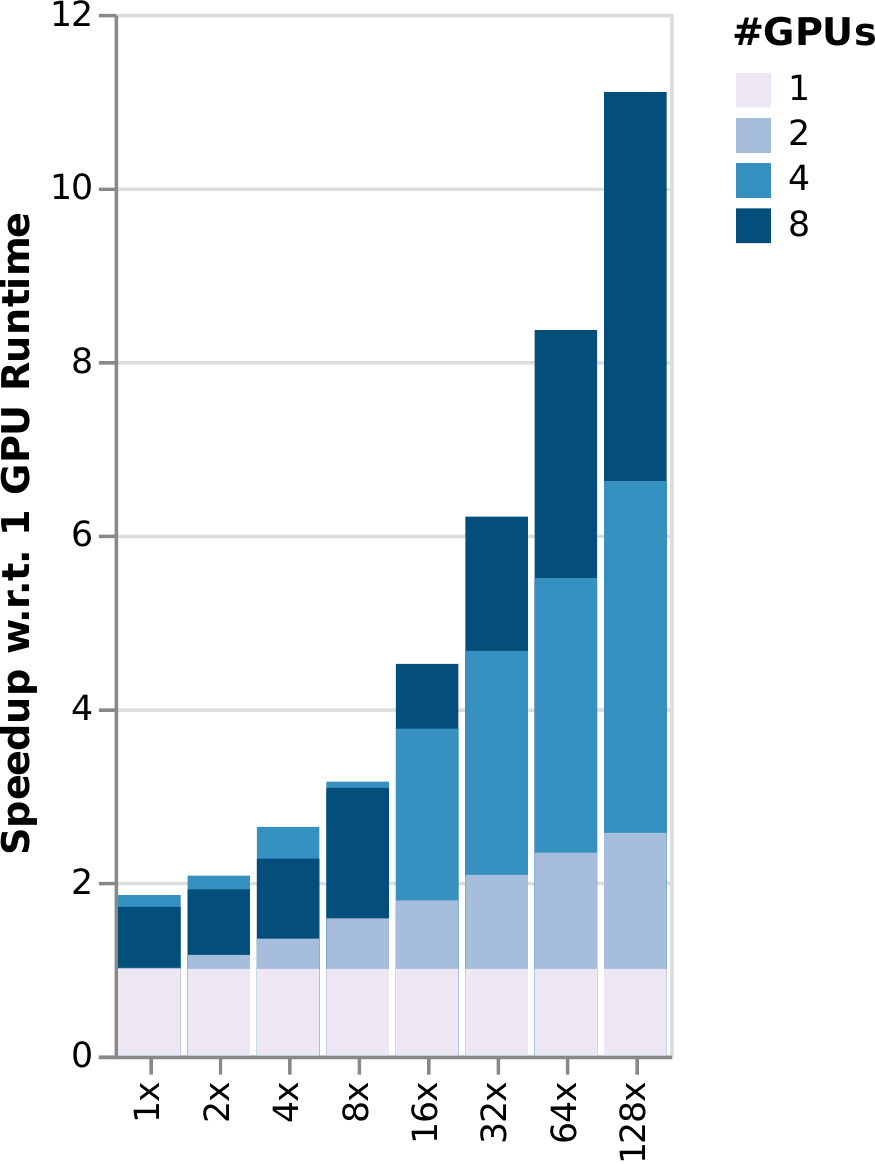}
    \caption{Speedup w.r.t. MG-GCN 1 GPU Runtime}
    \label{fig:density_speedup}
\end{figure}

The runtime of SpMM can be mainly divided into two parts: computation time and
communication time. Since we mostly overlap the two, the runtime can
be at best maximum of those two. Communication time only depends on the
dimensions of the matrix, whereas the computation time also depends on density
and sparsity structure of the matrix. Furthermore, computation time starts to
dominate the execution time as average degree increases. To illustrate the
effect of this on speedup, we used the synthetic datasets generated by scaling
the Arxiv dataset as explained in Section~\ref{sec:expr:setup}.
Figure~\ref{fig:density_speedup} displays the speedups obtained by $2$ to $8$ GPUs,
while we increase the average degree. As seen in the figure, our code
starts to achieve super-linear speedup with $2$ and $4$ GPUs, after $32 \times$, and with $8$ GPUs,
after $64 \times$ scaling. We attribute this super-linear speedup numbers for very
dense adjacency matrices because of the blocking effect of partitioning and
potentially better use of the cache.

\subsection{Comparison on Single Node Systems}

\begin{figure}[h]
    \centering
    \includegraphics[width=0.95\columnwidth]{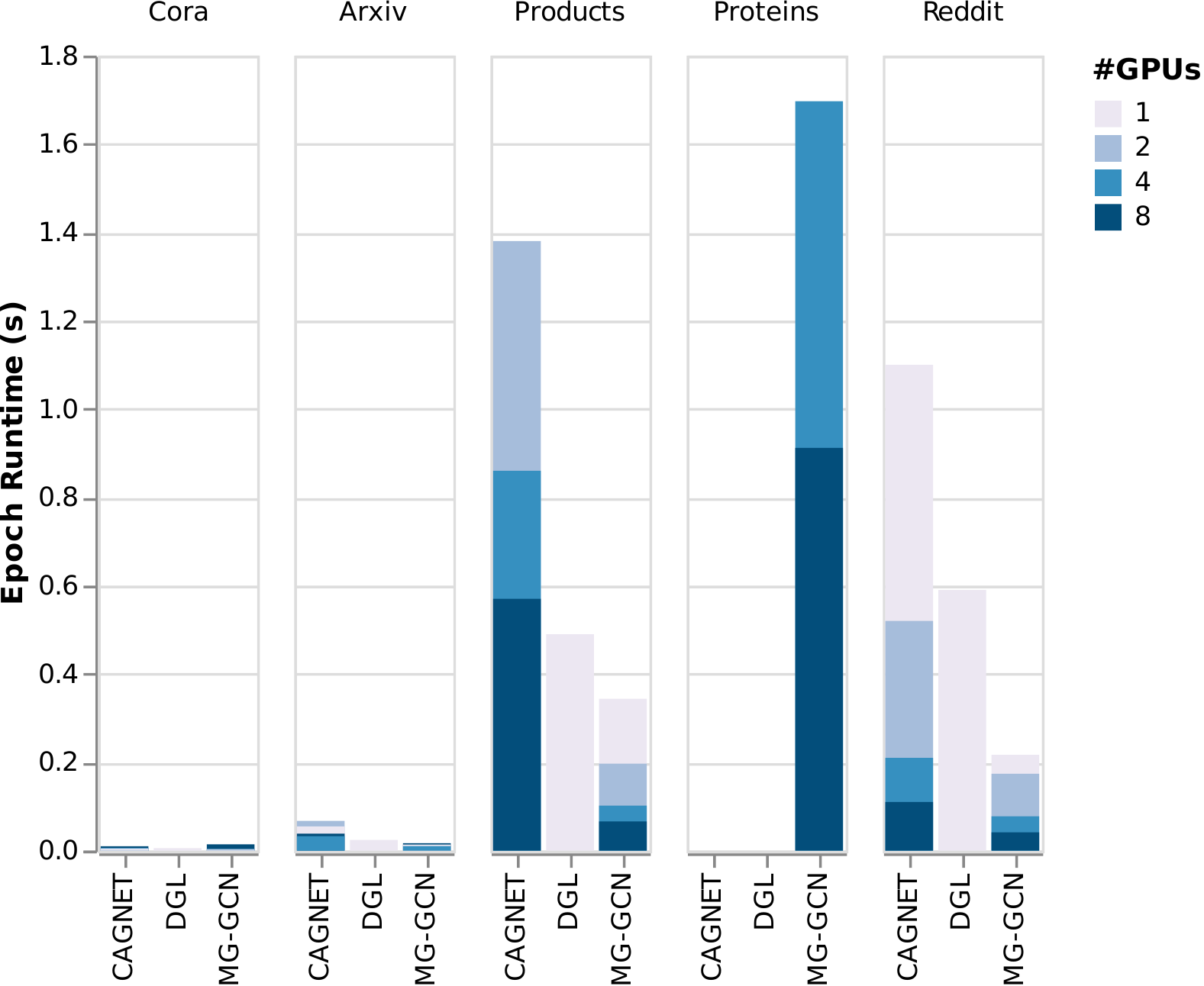}
    \caption{Baseline epoch runtime (seconds) comparison on DGX-V100. On Proteins dataset CAGNET and DGL run out of memory, MG-GCN runs out of memory with 1 and 2 GPUs.}
    \label{fig:runtime_baseline_v100}
\end{figure}

\begin{figure}[h]
    \centering
    \includegraphics[width=0.60\columnwidth]{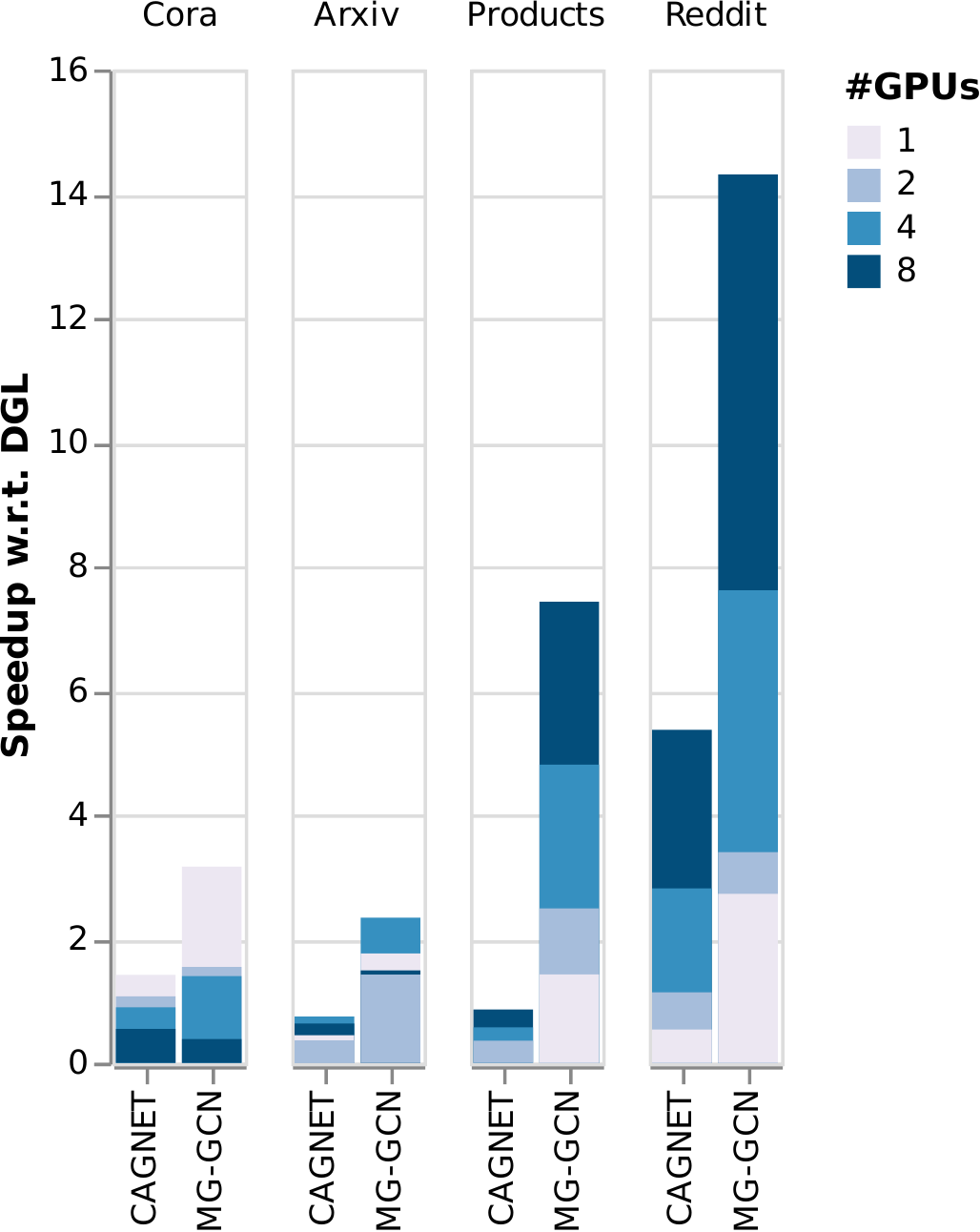}
    \caption{Speedup w.r.t. DGL on DGX-V100.}
    \label{fig:speedup_baseline_v100}
\end{figure}

\noindent{\em Comparison on DGX-V100:}
In Figures~\ref{fig:runtime_baseline_v100} and~\ref{fig:speedup_baseline_v100},
we compare MG-GCN with DGL and CAGNET using the 2 layer model mentioned
in Section~\ref{sec:Model} on DGX-V100. Note that, CAGNET has different partitioning strategies
namely, 1D, 1.5D, 2D and 3D. We present the best results which are produced by 1D partitioning.
In all datasets, we outperform DGL with a single GPU and CAGNET with multiple GPUs.
Our single GPU performances are, $2.72 \times$ faster on Reddit, $1.42 \times$ faster on Products, $1.76 \times$
faster on Arxiv and $3.1$x faster on Cora than DGL. Our 8 GPU performances are $2.66$x faster
on Reddit, $8.6 \times$ faster on Products, $2.35 \times$ faster on Arxiv than CAGNET.
Notice that, neither MG-GCN or CAGNET is able to get a speedup on Cora dataset,
since the graph is very small and certain amount of work is expected to achieve
any speedup. We are not able to run CAGNET with Proteins dataset
using 8 GPUs because of CAGNET's memory requirement; however, MG-GCN is
able to fit Proteins dataset into 4 only GPUs.
Even though, both CAGNET and MG-GCN use 1D partitioning,
we are able to fit much larger graphs into our target machines
due to extensive memory optimization described in Section~\ref{sec:memory_optim}.
Also, by overlapping computation and
communication, we achieve substantial speedup compared to CAGNET.

\begin{figure}[h]
    \centering
    \includegraphics[width=0.72\columnwidth]{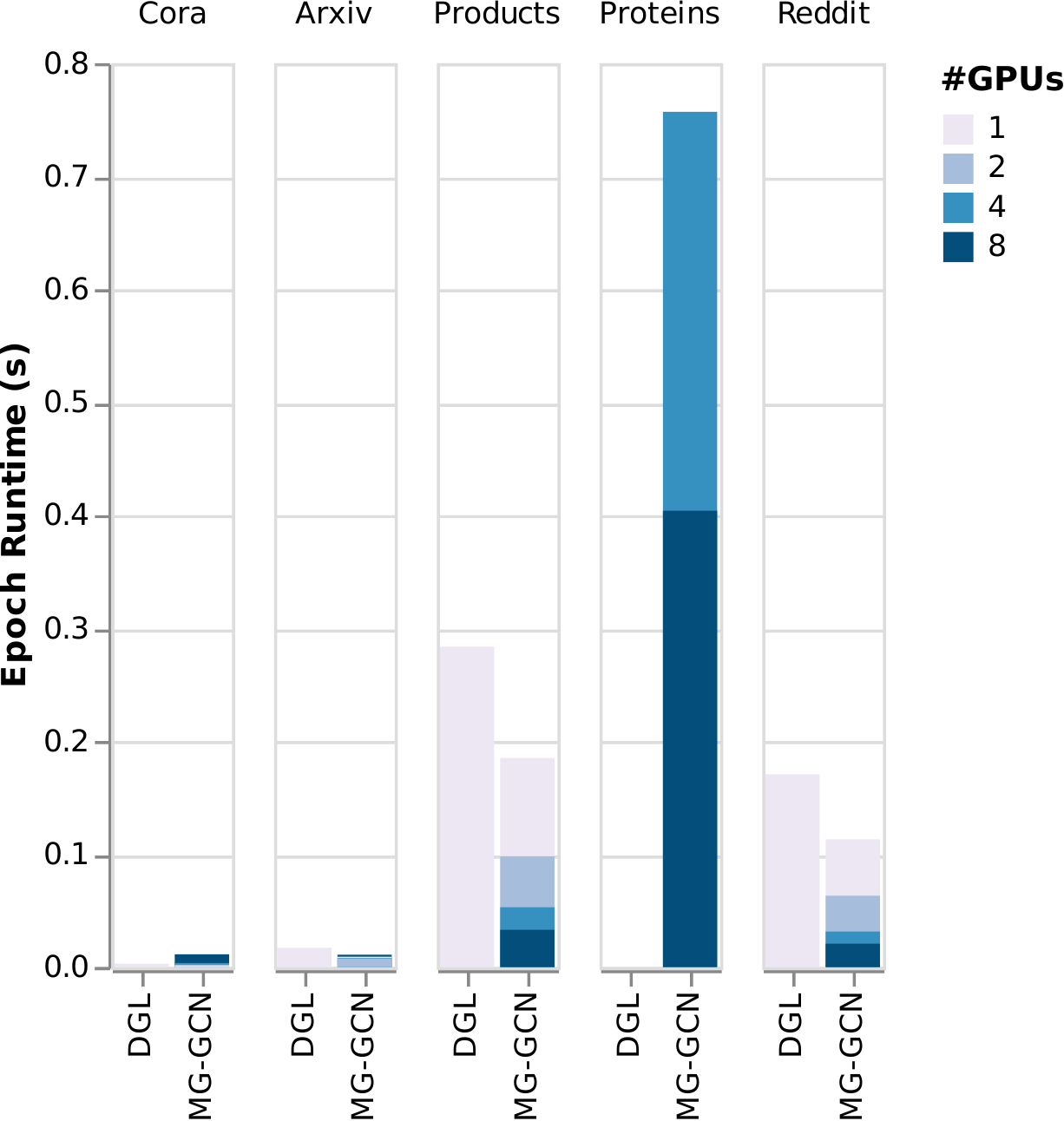}
    \caption{Epoch runtime (seconds) comparison on DGX-A100.}
    \label{fig:runtime_baseline_a100}
\end{figure}

\begin{figure}[h]
    \centering
    \includegraphics[width=0.95\columnwidth]{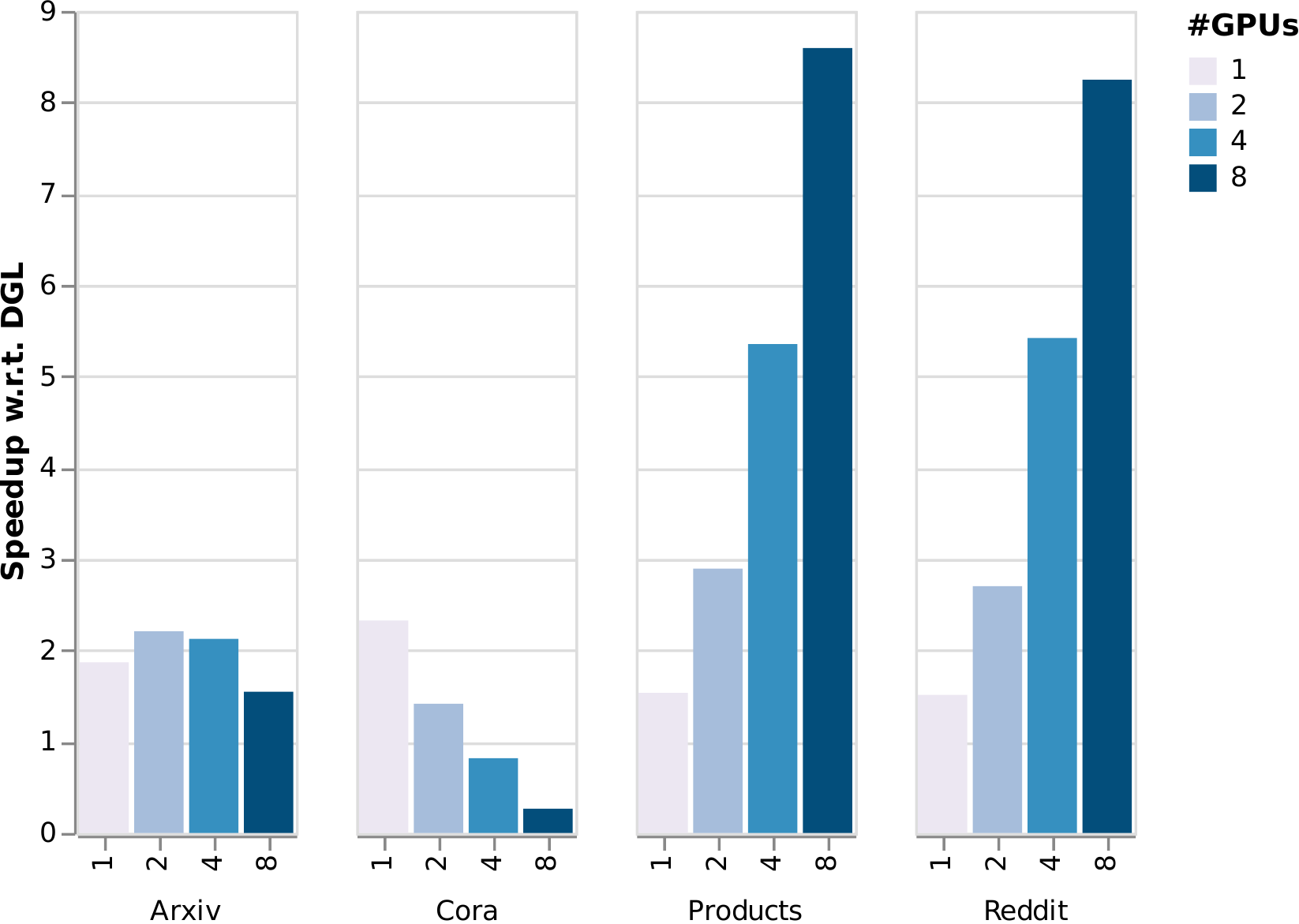}
    \caption{Speedup w.r.t. DGL on DGX-A100.}
    \label{fig:speedup_baseline_a100}
\end{figure}

\noindent{\em Comparison on DGX-A100:}
In Figures~\ref{fig:runtime_baseline_a100} and~\ref{fig:speedup_baseline_a100}, we compare
MG-GCN with DGL using the 2 layer GCN model mentioned in Section~\ref{sec:Model} on DGX-A100.
We are not able to include CAGNET in this comparison, since it is not compatible with CUDA 11.
In all the datasets, we outperform DGL with a single GPU. Our single GPU results are
$2.2 \times$ faster on Cora, $1.8 \times$ faster on Arxiv, $1.5 \times$ faster on Products and $1.5 \times$ faster on Reddit
datasets than DGL. On multi-GPU setting, MG-GCN is able to achieve $8.5 \times$ speedup on Products dataset,
and $8.3 \times$ speedup on Reddit dataset using 8 GPUs. Moreover, we are able to fit
Papers dataset, which is the largest available benchmark dataset for GNN training,
into 8 GPUs with MG-GCN, and achieve 2.89 seconds epoch runtime using the 4th
GCN model mentioned in Section~\ref{sec:Model}.

\subsection{Single Node vs Distributed Systems}

We compare MG-GCN with DistGNN using $2$ different GCN models mentioned in section~\ref{sec:Model}.
Note that, this is not an exact comparison for two main reasons: First, we are not able
reproduce the results because the source code of DistGNN is not available, so we base our
comparison to the numbers reported in the original work. Second, DistGNN is a CPU based framework,
while MG-GCN is designed for GPUs. We believe that, comparing the two frameworks will provide
important insights on the resource requirements and performance one can get.
For the experiments, DistGNN uses a cluster with 64 Intel Xeon 9242
CPU @2.30 GHz with 48 cores per socket in a dual-socket system. The compute nodes consist
of 384 GB memory, and connected through Mellanox HDR interconnect with DragonFly topology.
In addition, to run Papers on a single socket, they use a single-socket machine with 1.5TB memory.

Table~\ref{tab:DistGNN} shows the results from DistGNN, while Table~\ref{tab:MG-GCN}
shows the performance of MG-GCN on DGX-A100. In Table~\ref{tab:DistGNN},
we only take the single socket and the best socket performances for
each dataset from the original work~\cite{md2021distgnn}.
Also, note that, we compare against their baseline version, since other variants are
not exact computations but approximations. For detailed results, we refer interested readers
to~\cite{md2021distgnn}. Even though, the authors observe significant speedups in their experiments,
MG-GCN outperforms their best performance with a single GPU on all datasets except Proteins.
Our 8 GPU performances are $40 \times$ faster on Reddit, $12.6 \times$ faster on Papers, $12.4 \times$  faster on
Products, and $1.77 \times$ faster on Protein datasets than DistGNN's best
performances. Note that, for
Reddit dataset, since the GCN model is very small, $2$ layers with $16$ neurons, MG-GCN cannot
achieve speedup after 4 GPUs.

\begin{table}[h]
    \caption{\textbf{DistGNN Results:} The numbers in the cells are epoch times in second. For each dataset, we take results for 1 Socket and the number of sockets that performs the best from~\cite{DistDGL}. DS: Dataset, \#S: Number of Sockets.}
    \label{tab:DistGNN}
    \vskip 0.15in
    \begin{center}
    \begin{small}
    \begin{sc}
    \begin{tabular}{crrrr}
    \toprule
    \diagbox{\#S}{DS} & Reddit & Papers & Products & Protein\\
    \midrule
    1 & 0.60 & 1000 & 11 & 100\\
    16 & 0.61 & - & - & -\\
    64 & - & - & 1.74 & 2.63\\
    128 & - & 36.45 & - & -\\
    \bottomrule
    \end{tabular}
    \end{sc}
    \end{small}
    \end{center}
    \vskip -0.1in
\end{table}

\begin{table}[h]
    \caption{\textbf{MG-GCN Results on DGX-A100:} The values in the cells are epoch times in seconds. Dashed line represents configurations that run out of memory. DS: Dataset, \#G: Number of GPUs.}
    \label{tab:MG-GCN}
    \vskip 0.15in
    \begin{center}
    \begin{small}
    \begin{sc}
    \begin{tabular}{crrrr}
    \toprule
    \diagbox{\#G}{DS} & Reddit & Papers & Products & Protein\\
    \midrule
    1 & 0.033 & - & 0.355 & 4.221 \\
    2 & 0.017 & - & 0.202 & 2.272 \\
    4 & 0.012 & - & 0.110 & 1.191\\
    8 & 0.012 & 2.89 & 0.067 & 0.641\\
    \bottomrule
    \end{tabular}
    \end{sc}
    \end{small}
    \end{center}
    \vskip -0.1in
\end{table}

\section{Conclusion}

In this paper, we present MG-GCN, a single node multi-GPU GCN training framework which enables
efficient distributed training of GCNs over the full-graph. MG-GCN adapts a 1D row partitioning
strategy. It also adapts extensive memory optimizations by re-using/sharing the allocated buffers
across layers and forward/backward phases, and enables overlapping communication
and computation. We have demonstrated that, MG-GCN is able
to achieve significant runtime improvements over the available state-of-the-art
frameworks on single GPU systems. Moreover, going into the multi-GPU setting,
we are able to fit much larger graphs into  the memory of our target machines. 
In our single GPU experiments,
we achieve up-to $2.72 \times$ speedup compared to DGL on Reddit dataset,
and on multi-GPU experiments we achieve up-to $8.6 \times$ speedup on Products dataset compared
to CAGNET on DGX-V100.

In future work, we are aiming to extend our framework to multi-GPU clusters.
By doing so, we aim to train larger datasets and enable distributed
training of even larger scale GNNs.
Another future direction is to accelerate the
Sampled Dense Dense Matrix Multiplication (SDDMM) kernel to enable parallel
training of a number of other models such as Graph Attention Networks~\cite{GAT2018}.

\section{Acknowledgements}

We thank Prof. Polo Chau for providing us access to their DGX-A100 for our experiments.
This work was partially supported by the NSF grant CCF-1919021.

\bibliographystyle{plain}
\bibliography{project}

\begin{thebibliography}{10}

\bibitem{tensorflow}
Martín Abadi, Ashish Agarwal, Paul Barham, Eugene Brevdo, Zhifeng Chen, Craig
  Citro, Greg~S. Corrado, Andy Davis, Jeffrey Dean, Matthieu Devin, Sanjay
  Ghemawat, Ian Goodfellow, Andrew Harp, Geoffrey Irving, Michael Isard, Rafal
  Jozefowicz, Yangqing Jia, Lukasz Kaiser, Manjunath Kudlur, Josh Levenberg,
  Dan Mané, Mike Schuster, Rajat Monga, Sherry Moore, Derek Murray, Chris
  Olah, Jonathon Shlens, Benoit Steiner, Ilya Sutskever, Kunal Talwar, Paul
  Tucker, Vincent Vanhoucke, Vijay Vasudevan, Fernanda Viégas, Oriol Vinyals,
  Pete Warden, Martin Wattenberg, Martin Wicke, Yuan Yu, and Xiaoqiang Zheng.
\newblock Tensorflow, October 2021.

\bibitem{ben2019demystifying}
Tal Ben-Nun and Torsten Hoefler.
\newblock Demystifying parallel and distributed deep learning: An in-depth
  concurrency analysis.
\newblock {\em ACM Computing Surveys (CSUR)}, 52(4):1--43, 2019.

\bibitem{chen2018fastgcn}
Jie Chen, Tengfei Ma, and Cao Xiao.
\newblock Fast{GCN}: Fast learning with graph convolutional networks via
  importance sampling.
\newblock In {\em International Conference on Learning Representations}, 2018.

\bibitem{chen2015mxnet}
Tianqi Chen, Mu~Li, Yutian Li, Min Lin, Naiyan Wang, Minjie Wang, Tianjun Xiao,
  Bing Xu, Chiyuan Zhang, and Zheng Zhang.
\newblock Mxnet: A flexible and efficient machine learning library for
  heterogeneous distributed systems, 2015.

\bibitem{chiang2019cluster}
Wei-Lin Chiang, Xuanqing Liu, Si~Si, Yang Li, Samy Bengio, and Cho-Jui Hsieh.
\newblock Cluster-gcn: An efficient algorithm for training deep and large graph
  convolutional networks.
\newblock In {\em Proceedings of the 25th ACM SIGKDD International Conference
  on Knowledge Discovery \& Data Mining}, pages 257--266, 2019.

\bibitem{coates2013deep}
Adam Coates, Brody Huval, Tao Wang, David Wu, Bryan Catanzaro, and Ng~Andrew.
\newblock Deep learning with cots hpc systems.
\newblock In {\em International conference on machine learning}, pages
  1337--1345. PMLR, 2013.

\bibitem{logistic}
D.~R. Cox.
\newblock The regression analysis of binary sequences.
\newblock {\em Journal of the Royal Statistical Society. Series B
  (Methodological)}, 20(2):215--242, 1958.

\bibitem{scalablestacking}
Li~Deng, Dong Yu, and John Platt.
\newblock Scalable stacking and learning for building deep architectures.
\newblock In {\em 2012 IEEE International Conference on Acoustics, Speech and
  Signal Processing (ICASSP)}, pages 2133--2136, 2012.

\bibitem{ericson2017performance}
Ludvig Ericson and Rendani Mbuvha.
\newblock On the performance of network parallel training in artificial neural
  networks.
\newblock {\em arXiv preprint arXiv:1701.05130}, 2017.

\bibitem{pyg}
Matthias Fey and Jan~E. Lenssen.
\newblock Fast graph representation learning with {PyTorch Geometric}.
\newblock In {\em ICLR Workshop on Representation Learning on Graphs and
  Manifolds}, 2019.

\bibitem{Gabert21-GrAPL}
Kasimir Gabert and \"Umit~V. \c{C}ataly\"{u}rek.
\newblock {PIGO}: A parallel graph input/output library.
\newblock In {\em 2021 IEEE International Parallel and Distributed Processing
  Symposium Workshops (IPDPSW)}, pages 276--279. IEEE, 2021.

\bibitem{ginsburg2018large}
Boris Ginsburg, Igor Gitman, and Yang You.
\newblock Large batch training of convolutional networks with layer-wise
  adaptive rate scaling, 2018.

\bibitem{powergraph}
Joseph~E. Gonzalez, Yucheng Low, Haijie Gu, Danny Bickson, and Carlos Guestrin.
\newblock Powergraph: Distributed graph-parallel computation on natural graphs.
\newblock In {\em Proceedings of the 10th USENIX Conference on Operating
  Systems Design and Implementation}, OSDI'12, page 17–30. USENIX
  Association, 2012.

\bibitem{goyal2017accurate}
Priya Goyal, Piotr Doll{\'a}r, Ross Girshick, Pieter Noordhuis, Lukasz
  Wesolowski, Aapo Kyrola, Andrew Tulloch, Yangqing Jia, and Kaiming He.
\newblock Accurate, large minibatch sgd: Training imagenet in 1 hour.
\newblock {\em arXiv preprint arXiv:1706.02677}, 2017.

\bibitem{hamilton2018inductive}
William~L. Hamilton, Rex Ying, and Jure Leskovec.
\newblock Inductive representation learning on large graphs, 2018.

\bibitem{hu2021open}
Weihua Hu, Matthias Fey, Marinka Zitnik, Yuxiao Dong, Hongyu Ren, Bowen Liu,
  Michele Catasta, and Jure Leskovec.
\newblock Open graph benchmark: Datasets for machine learning on graphs, 2021.

\bibitem{ROC}
Zhihao Jia, Sina Lin, Mingyu Gao, Matei Zaharia, and Alex Aiken.
\newblock Improving the accuracy, scalability, and performance of graph neural
  networks with roc.
\newblock {\em Proceedings of Machine Learning and Systems (MLSys)}, pages
  187--198, 2020.

\bibitem{spmm}
Peng Jiang, Changwan Hong, and Gagan Agrawal.
\newblock A novel data transformation and execution strategy for accelerating
  sparse matrix multiplication on {GPU}s.
\newblock In {\em Proceedings of the 25th ACM SIGPLAN Symposium on Principles
  and Practice of Parallel Programming}, PPoPP '20, page 376–388. Association
  for Computing Machinery, 2020.

\bibitem{metis}
George Karypis and Vipin Kumar.
\newblock A fast and high quality multilevel scheme for partitioning irregular
  graphs.
\newblock {\em SIAM Journal on Scientific Computing}, 20:359--392, 1998.

\bibitem{kingma2017adam}
Diederik~P. Kingma and Jimmy Ba.
\newblock Adam: A method for stochastic optimization, 2017.

\bibitem{gcn}
Thomas~N. Kipf and Max Welling.
\newblock Semi-supervised classification with graph convolutional networks.
\newblock In {\em 5th International Conference on Learning Representations,
  {ICLR} 2017, Toulon, France, April 24-26, 2017, Conference Track
  Proceedings}. OpenReview.net, 2017.

\bibitem{Kolda_2014}
Tamara~G. Kolda, Ali Pinar, Todd Plantenga, and C.~Seshadhri.
\newblock A scalable generative graph model with community structure.
\newblock {\em SIAM Journal on Scientific Computing}, 36(5):C424–C452, Jan
  2014.

\bibitem{krizhevsky2014one}
Alex Krizhevsky.
\newblock One weird trick for parallelizing convolutional neural networks.
\newblock {\em arXiv preprint arXiv:1404.5997}, 2014.

\bibitem{lenet}
Y.~Lecun, L.~Bottou, Y.~Bengio, and P.~Haffner.
\newblock Gradient-based learning applied to document recognition.
\newblock {\em Proceedings of the IEEE}, 86(11):2278--2324, 1998.

\bibitem{dgx}
Ang Li, Shuaiwen~Leon Song, Jieyang Chen, Jiajia Li, Xu~Liu, Nathan~R. Tallent,
  and Kevin~J. Barker.
\newblock Evaluating modern {GPU} interconnect: {PCIe}, {NVLink}, {NV-SLI},
  {NVSwitch} and {GPUDirect}.
\newblock {\em IEEE Transactions on Parallel and Distributed Systems},
  31(1):94–110, Jan 2020.

\bibitem{neugraph}
Lingxiao Ma, Zhi Yang, Youshan Miao, Jilong Xue, Ming Wu, Lidong Zhou, and
  Yafei Dai.
\newblock Neugraph: Parallel deep neural network computation on large graphs.
\newblock In {\em 2019 {USENIX} Annual Technical Conference ({USENIX} {ATC}
  19)}, pages 443--458. {USENIX} Association, July 2019.

\bibitem{md2021distgnn}
Vasimuddin Md, Sanchit Misra, Guixiang Ma, Ramanarayan Mohanty, Evangelos
  Georganas, Alexander Heinecke, Dhiraj Kalamkar, Nesreen~K Ahmed, and
  Sasikanth Avancha.
\newblock Distgnn: Scalable distributed training for large-scale graph neural
  networks.
\newblock {\em arXiv preprint arXiv:2104.06700}, 2021.

\bibitem{relu}
Vinod Nair and Geoffrey~E. Hinton.
\newblock Rectified linear units improve restricted boltzmann machines.
\newblock In {\em Proceedings of the 27th International Conference on
  International Conference on Machine Learning}, ICML'10, page 807–814,
  Madison, WI, USA, 2010. Omnipress.

\bibitem{pytorch}
Adam Paszke, Sam Gross, Francisco Massa, Adam Lerer, James Bradbury, Gregory
  Chanan, Trevor Killeen, Zeming Lin, Natalia Gimelshein, Luca Antiga, Alban
  Desmaison, Andreas Kopf, Edward Yang, Zachary DeVito, Martin Raison, Alykhan
  Tejani, Sasank Chilamkurthy, Benoit Steiner, Lu~Fang, Junjie Bai, and Soumith
  Chintala.
\newblock Pytorch: An imperative style, high-performance deep learning library.
\newblock In H.~Wallach, H.~Larochelle, A.~Beygelzimer, F.~d\textquotesingle
  Alch\'{e}-Buc, E.~Fox, and R.~Garnett, editors, {\em Advances in Neural
  Information Processing Systems 32}, pages 8024--8035. Curran Associates,
  Inc., 2019.

\bibitem{GNN}
Franco Scarselli, Marco Gori, Ah~Chung Tsoi, Markus Hagenbuchner, and Gabriele
  Monfardini.
\newblock The graph neural network model.
\newblock {\em IEEE Transactions on Neural Networks}, 20(1):61--80, 2009.

\bibitem{sen:aimag08}
Prithviraj Sen, Galileo~Mark Namata, Mustafa Bilgic, Lise Getoor, Brian
  Gallagher, and Tina Eliassi-Rad.
\newblock Collective classification in network data.
\newblock {\em AI Magazine}, 29(3):93--106, 2008.

\bibitem{CAGNET}
Alok Tripathy, Katherine~A. Yelick, and Aydin Bulu{\c{c}}.
\newblock Reducing communication in graph neural network training.
\newblock {\em CoRR}, abs/2005.03300, 2020.

\bibitem{Geijn1997SUMMASU}
Robert~A. van~de Geijn and Jerrell Watts.
\newblock Summa: scalable universal matrix multiplication algorithm.
\newblock {\em Concurr. Pract. Exp.}, 9:255--274, 1997.

\bibitem{GAT2018}
Petar Veličković, Guillem Cucurull, Arantxa Casanova, Adriana Romero, Pietro
  Liò, and Yoshua Bengio.
\newblock Graph attention networks, 2018.

\bibitem{Wang2019DeepGL}
Minjie Wang, Da~Zheng, Zi-Hao Ye, Q.~Gan, Mufei Li, Xiang Song, Jinjing Zhou,
  Chao Ma, Lingfan Yu, Yu~Gai, Tianjun Xiao, Tong He, G.~Karypis, Jinyang Li,
  and Zheng Zhang.
\newblock Deep graph library: A graph-centric, highly-performant package for
  graph neural networks.
\newblock 2019.

\bibitem{wang2019dgl}
Minjie Wang, Da~Zheng, Zihao Ye, Quan Gan, Mufei Li, Xiang Song, Jinjing Zhou,
  Chao Ma, Lingfan Yu, Yu~Gai, Tianjun Xiao, Tong He, George Karypis, Jinyang
  Li, and Zheng Zhang.
\newblock Deep graph library: A graph-centric, highly-performant package for
  graph neural networks.
\newblock {\em arXiv preprint arXiv:1909.01315}, 2019.

\bibitem{Libra}
Cong Xie, Ling Yan, Wu-Jun Li, and Zhihua Zhang.
\newblock Distributed power-law graph computing: Theoretical and empirical
  analysis.
\newblock In {\em Advances in Neural Information Processing Systems},
  volume~27. Curran Associates, Inc., 2014.

\bibitem{Zhang2014}
Kunlei Zhang and Xue-Wen Chen.
\newblock Large-scale deep belief nets with mapreduce.
\newblock {\em IEEE Access}, 2:395--403, 2014.

\bibitem{zhang2018link}
Muhan Zhang and Yixin Chen.
\newblock Link prediction based on graph neural networks.
\newblock {\em Advances in Neural Information Processing Systems},
  31:5165--5175, 2018.

\bibitem{zhang2018end}
Muhan Zhang, Zhicheng Cui, Marion Neumann, and Yixin Chen.
\newblock An end-to-end deep learning architecture for graph classification.
\newblock In {\em Thirty-Second AAAI Conference on Artificial Intelligence},
  2018.

\bibitem{DistDGL}
Da~Zheng, Chao Ma, Minjie Wang, Jinjing Zhou, Qidong Su, Xiang Song, Quan Gan,
  Zheng Zhang, and George Karypis.
\newblock Distdgl: distributed graph neural network training for billion-scale
  graphs.
\newblock In {\em 2020 IEEE/ACM 10th Workshop on Irregular Applications:
  Architectures and Algorithms (IA3)}, pages 36--44. IEEE, 2020.

\bibitem{Aligraph}
Rong Zhu, Kun Zhao, Hongxia Yang, Wei Lin, Chang Zhou, Baole Ai, Yong Li, and
  Jingren Zhou.
\newblock Aligraph: A comprehensive graph neural network platform.
\newblock {\em Proc. VLDB Endow.}, 12(12):2094–2105, August 2019.

\bibitem{zinkevich2010parallelized}
Martin Zinkevich, Markus Weimer, Alexander~J Smola, and Lihong Li.
\newblock Parallelized stochastic gradient descent.
\newblock In {\em NIPS}, volume~4, page~4. Citeseer, 2010.

\end{thebibliography}

\end{document}